\documentclass{article} %
\usepackage{iclr2026_conference,times}

\usepackage{amsmath,amsfonts,bm}

\def\eqref#1{equation~\ref{#1}}

\def\1{\bm{1}}

\DeclareMathAlphabet{\mathsfit}{\encodingdefault}{\sfdefault}{m}{sl}
\SetMathAlphabet{\mathsfit}{bold}{\encodingdefault}{\sfdefault}{bx}{n}

\usepackage{hyperref}
\usepackage{url}
\usepackage{booktabs}
\usepackage{multirow}
\usepackage{xcolor}
\usepackage{colortbl}
\usepackage{threeparttable}
\usepackage[utf8]{inputenc}
\usepackage[T1]{fontenc}

\usepackage[ampersand]{easylist}
\ListProperties(Hide2=1,Hide3=2,Progressive*=.5cm,Numbers3=l, Numbers4=r,FinalMark3={)})
\usepackage{etoolbox}
\AtBeginEnvironment{easylist}{\ListProperties(Start1=1)}

\usepackage{xcolor}

\definecolor{cblue}{RGB}{8, 85, 153}
\usepackage{tcolorbox}
\tcbuselibrary{skins}
\usepackage{makecell}
\usepackage{fontawesome5} %
\definecolor{boxblue}{RGB}{59, 130, 246}
\definecolor{boxbluelight}{RGB}{239, 246, 255}
\definecolor{boxgreen}{RGB}{34, 197, 94}
\definecolor{boxgreenlight}{RGB}{240, 253, 244}
\definecolor{boxpurple}{RGB}{147, 51, 234}
\definecolor{boxpurplelight}{RGB}{250, 245, 255}
\definecolor{boxorange}{RGB}{249, 115, 22}
\definecolor{boxorangelight}{RGB}{255, 247, 237}
\definecolor{boxred}{RGB}{239, 68, 68}
\definecolor{boxredlight}{RGB}{254, 242, 242}

\usepackage[capitalise]{cleveref}
\usepackage{float}
\usepackage{placeins} %
\usepackage{adjustbox}

\usepackage{makecell}
\usepackage[ampersand]{easylist}
\usepackage{todonotes}
\usepackage{caption}
\usepackage{xspace}
\usepackage{markdown}
\usepackage{multirow}
\usepackage{array}
\usepackage{booktabs}
\usepackage{tabularx}
\usepackage{graphicx}
\usepackage{ulem}
\usepackage{enumitem}

\newcolumntype{Y}{>{\centering\arraybackslash}X}

\newtcolorbox{custombox}[2][blue]{
  enhanced,
  colback={box#1light},
  colframe={box#1},
  coltitle=white,
  colbacktitle={box#1},
  fonttitle={\bfseries\large},
  title={#2},
  rounded corners,
  boxrule=2pt,
  left=10pt,
  right=10pt,
  top=10pt,
  bottom=10pt
}
\newcommand{\methodlong}{Ensemble explanations\xspace}

\usepackage{adjustbox}
\usepackage{listings}
\usepackage{tikz} %

\title{Do explanations generalize across large reasoning models?}

\author{
Koyena Pal$^{1}$\thanks{Work done as part of the CBAI Fellowship},\;\; David Bau$^{1}$,\; Chandan Singh$^{2}$ \\
$^{1}$Northeastern University \quad $^{2}$Microsoft Research \\
\texttt{\{pal.k, d.bau\}@northeastern.edu, chansingh@microsoft.com}
}

\iclrfinalcopy
\begin{document}
\maketitle
\begin{abstract}
Large reasoning models (LRMs) produce a textual chain of thought (CoT) in the process of solving a problem, which serves as a potentially powerful tool to understand the problem by surfacing a human-readable, natural-language explanation.
However, it is unclear whether these explanations \textit{generalize}, i.e. whether they capture general patterns about the underlying problem rather than patterns which are esoteric to the LRM.
This is a crucial question in
understanding or discovering new concepts, e.g. in AI for science.
We study this generalization question by evaluating a specific notion of generalizability:
whether explanations produced by one LRM induce the same behavior when given to other LRMs.
We find that CoT explanations often exhibit this form of generalization (i.e. they increase consistency between LRMs)
and that this increased generalization is correlated with human preference rankings and post-training with reinforcement learning.
We further analyze the conditions under which explanations yield consistent answers and propose a straightforward, sentence-level ensembling strategy that improves consistency.
Taken together, these results prescribe caution when using LRM explanations to yield new insights and outline a framework for characterizing LRM explanation generalization.
\end{abstract}

\section{Introduction}
The chains of thought (CoTs) produced by large reasoning models (LRMs) have enabled strong performance on a range of complex tasks~\citep{guo2025deepseek,guha2025openthoughts,liu2025prorl,abdin2025phi,agarwal2025gpt}.
These CoTs are often presented as human-readable explanations, but many researchers have questioned whether these traces can be made \emph{faithful} to the true decision-making processes followed by LRMs~\citep{barez2025chain,chen2023models,shojaee2025illusion,xiong2025measuring}. Another perspective of understanding the model's chain-of-thought can be in terms of its utility as an explanation, i.e., not just whether it reflects the model's internal reasoning, but whether it can effectively communicate that reasoning to other models and humans. Hence, in this paper, we examine a different property that is related to faithfulness: we investigate the \emph{generalization} of reasoning traces across different LRMs.

Our study is motivated by the search for good natural-language explanations.
For an explanation to be useful, it must not only be accurate, but also \textit{understandable}; i.e. when presented to a new person (or LLM), the explanation should lead them to draw the intended conclusions from it.
In our setting, we quantify this question in terms of cross-LRM CoT generalization.
Specifically, we ask whether an explanation produced from one LRM reliably guides other LRMs to the same answer. 

Posing the evaluation in this way enables an automated, quantitative evaluation of generalization for explanations, which has remained elusive despite generalization being a cornerstone of statistical machine learning.
This evaluation is especially critical in scientific discovery, where explanations that capture problem-level patterns, rather than model-specific quirks,
could inspire novel human insights~\citep{schut2025bridging,singh2024rethinking}, especially as LRMs reach superhuman capabilities in domains such as science/mathematics~\citep{wang2023scientific,romera2024mathematical} and are increasingly used in educational settings~\citep{kasneci2023chatgpt,bewersdorff2025taking}.

\begin{figure}[!ht]
    \centering
    \includegraphics[width=\linewidth]{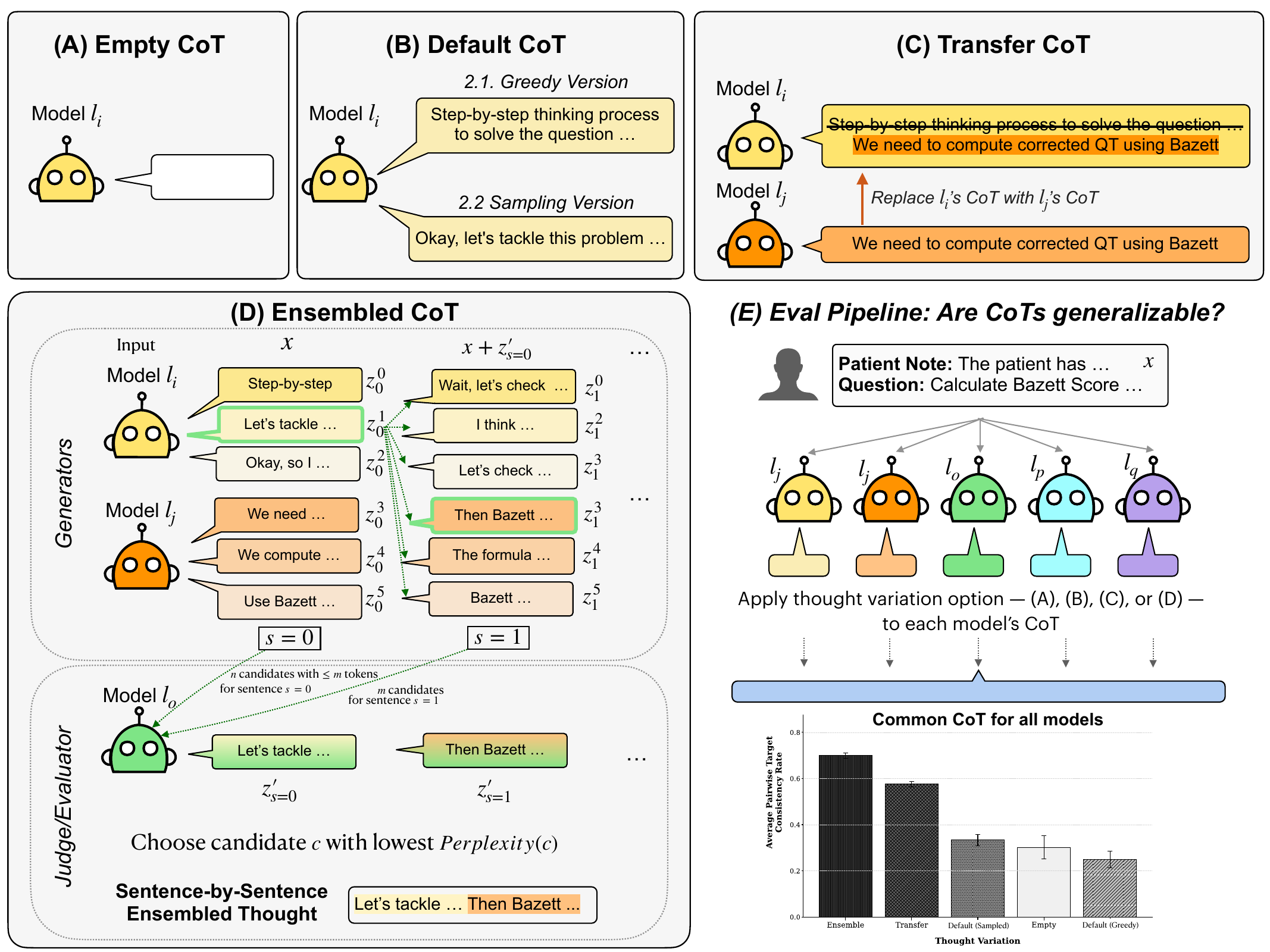}
    \caption{\textit{Methods for eliciting chain-of-thought (CoT) explanations and evaluating them.}
We evaluate explanation generation by querying LLMs for the answer to questions using CoTs in four different ways:
\textbf{(A) Empty CoT.} No reasoning text is provided between the model’s thinking tags.
\textbf{(B) Default CoT.} The model’s own reasoning is used. (2.1) uses greedy decoding, while (2.2) uses nucleus sampling.
\textbf{(C) Transfer CoT.} Reasoning from one model is directly transferred to another, replacing its own.
\textbf{(D) Ensembled CoT.} A generator–evaluator loop. Generator models produce $n=3$ candidate sentences ($\leq 15$ tokens each), forming $k$ candidates. These are scored by the evaluator, and the least surprising candidate (lowest perplexity) is appended to the growing ensembled thought. This updated context is fed back into the generators, and the process repeats until an end-of-thought or maximum token limit is reached.
\textbf{(E)} After eliciting CoTs in these four settings, we evaluate their generalization to new LRMs.
The transfer CoT and Ensembled CoT significantly improve the consistency of the answer produced between the model producing the original CoT and the model producing the final answer (results shown here on the MedCalc-Bench dataset).
}
    \label{fig:overview}
\end{figure}

We evaluate the effect of LRM explanations on improving the consistency between LRM answers, when the explanation CoTs are generated in different ways (\cref{fig:overview}A-C).
To further improve generalization, we propose a sentence-level ensembling strategy that encourages the production of explanations less tied to the idiosyncrasies of any single model and find that it increases consistency between LRMs (\cref{fig:overview}D).
We find that LRM explanations do generalize, i.e. they increase consistency between LRMs (\cref{fig:overview}E), even improving consistency when the underlying explanation suggests a wrong answer.
Moreover, the Ensembled CoT improves consistency more than other strategies for eliciting a CoT.

We then evaluate the relationship between cross-model consistency and two other aspects.
First, we conduct a human study evaluating human preferences for various CoT explanations, and find that more consistent explanations also appear preferable for human users.
Second, we evaluate the relationship between cross-model consistency and post-training with reinforcement learning (RL), and find that RL post-training yields CoTs that are more consistent with other LRMs.
Together, these results represent a step towards understanding when explanations from LRMs can be both transferable across models and informative to human users.

\section{Methods}\label{sec:methods}

\paragraph{Eliciting CoT explanations}
We seek to evaluate the generalization of a CoT explanation from one LRM to a new LRM.
\cref{fig:overview}A-D gives an overview of the different methods for eliciting LRM explanations that we consider.
Concretely, given an LRM $l_{\text{gen}}$ and a problem string $x$, we elicit a CoT explanation by passing a model-specific prompt that elicits an explanation $z = l_{\text{gen}}(x)$ before producing an answer $a = l_{\text{gen}}(x | z)$.
For example, with the \texttt{Qwen/QwQ-32B} model~\citep{qwq32b}, we use a prompt of the form:
\texttt{\{Problem\}<think>\{Thinking Text\}</think>\{Answer\}}
We evaluate four variations of CoT generation across models:
\begin{enumerate}[noitemsep,leftmargin=*]
    \item Empty CoT (\cref{fig:overview}A): The think text is an empty string, serving as a baseline method. Therefore, when the model generates its final answer, the preceding context is \texttt{\{Problem\}<starting-think-tag>""<closing-think-tag>}
    \item Default CoT (\cref{fig:overview}B): The standard setting used in prior benchmarks, where the think text is generated by the model without modification. This method includes two sub-variations: one without sampling and one with sampling, covering both deterministic and non-deterministic default behaviors.
    \item Transfer CoT (\cref{fig:overview}C): The think text is replaced by a default deterministic think text of another model. We test on various permutations of the models to see how different models' reasoning traces generalize across other models.
    \item Ensembled Thoughts (\cref{fig:overview}D): The think text is replaced by explanations generated via \methodlong. %
    Given a set of LRMs, we designate a subset as generators and a separate model as the evaluator. At each step, the generators produce $n$ candidate sentences with $m$ tokens ($n=3$ and $m=15$, in our case) conditioned on the context, which consists of the problem string $x$. The evaluator then selects the candidate with the lowest perplexity, which is appended to the ensembled CoT. The context is subsequently updated to include the original problem and all accumulated ensembled sentences. This sentence is part of $z$, which would be of size $s$, where size indicates the number of sentences we generate to create a complete ensembled thought. This process repeats until one of the generator models outputs an end-of-thought token or the maximum chain length ($m \cdot s$) is reached.
\end{enumerate}

\paragraph{Evaluating CoT explanations}
Given the explanation from $l_{\text{gen}}$, i.e., $z = l_{\text{gen}}(x)$, we then test its generalization to a new LRM $l_{\text{eval}}$ (see \cref{fig:overview}E).
Specifically, we construct a prompt that includes the explanation within \texttt{think} tags, and then use it to query $l_{\text{eval}}$ for an answer.

We measure two metrics: \textit{Accuracy}, which measures the match between the generated answer and the ground truth answer $a$, and \textit{Consistency}, which measures the match between answers from different models in a population of models $L = \{l_i,l_j ...l_q\}$ when given the same explanation.

\begin{align}
    \text{Accuracy} = I(l_{\text{eval}}(x|l_{\text{gen}}(x)),a), \quad \quad \text{Consistency} = \sum_{\substack{l \in L \\ z = l_{\text{gen}}(x)}}I(l_{\text{eval}}(x|z), l(x|z)) \label{eq:consistency}
\end{align}
where $I$ is the scoring function specific to a dataset to see if two answers match.

One potential issue with this method of testing generalization is that explanations may directly include an answer, rather than an explanation leading to an answer.
To avoid this issue, we process each explanation by prompting a separate LLM (gpt-o4-mini, \texttt{o4-mini-2025-04-16}, \citeauthor{openaio4mini}~\citeyear{openaio4mini}) to remove any explicit answer declarations detected in the explanation (see details in \cref{sec:remove_answer_method}).

\subsection{Experimental setup}

\paragraph{Datasets and evaluation}
To assess reasoning in both specialized and general domains, we leverage two benchmarks.
First, we use \texttt{MedCalc-Bench}~\citep{khandekar2024medcalc}, which targets medical domain-specific reasoning.
Each instance consists of a patient note and a question, which asks to compute a specific clinical value, and we randomly select 100 examples of the dataset for evaluation.
Second, we study \texttt{Instruction Induction}~\citep{honovich2022instruction}, which evaluates general reasoning capabilities.
Each instance presents five input-output pairs and the model is tasked to generate a natural language instruction that captures their underlying relation.
We randomly selected 8 of the original Instruction Induction tasks and extend the latter benchmark with 12 novel tasks to capture more complex general reasoning (see details in \cref{sec:instruction_induction_data}).
We evaluate the resulting 20 tasks by taking 100 random examples, five per task.

To evaluate results on these datasets, we specify  the scoring function $I$ from \cref{eq:consistency} to be exact-matching for \texttt{MedCalc-Bench} and BERTScore~\citep{zhang2019bertscore} for \texttt{Instruction Induction}.
We evaluate CoT generalization across the models in \cref{tab:model_info}.

\paragraph{User study}
To investigate whether greater generalizability correlates with users' perceptions of model CoT quality,
we designed and conducted a user study.
Participants were presented a CoT for a problem and asked to evaluate it
across the following criteria\footnote{Note that the questions ask users to evaluate the explanations' quality rather than directly use the explanation to perform a task; this limitation is largely because the MedCalc-Bench dataset requires significant domain expertise which users often did not have.}
:
\begin{itemize}[noitemsep,leftmargin=*]
    \item \textit{Clarity of Steps:} The reasoning steps were clear and well explained (1 = Very unclear; 5 = Very clear)
    \item \textit{Ease of Following}: The answer follows clearly from the reasoning steps. (1 = Very difficult; 5 = Very confident)
    \item \textit{Confidence:} After reading, I feel confident I understood the reasoning. (1 = Not confident at all; 5 = Very confident)
    \item \textit{Best Overall} ranking was asked in the following way: ``Rank the following models' Chain-of-Thought explanations from most understandable to least understandable" (1 is the most understandable)
\end{itemize}

The study used 10 problems on the MedCalc-Bench dataset. For each problem, a CoT was generated in 4 different ways (see \cref{tab:user_study_model_info}).
For evaluation, participants were shown 5 examples, randomly selected and balanced across conditions.
The study was administered via Qualtrics, where 15 participants, all of whom were computer science or healthcare researchers, received an anonymous survey link. 
Answers were given on a 5-point Likert scale.
See more design details in \cref{sec:user_study_details}.

\begin{table}[t]
\centering
\small
\caption{LRMs used in this study. We focus on recent LRMs that have shown strong performance in various reasoning tasks.}
{\small
\begin{tabular}{@{}l l p{5cm} p{2.5cm}@{}}
\toprule
\textbf{Alias} & \textbf{Model} & \textbf{Huggingface ID} & \textbf{Citation} \\
\midrule
NRR & \parbox{3.5cm}{Nemotron-Research-Reasoning-Qwen-1.5B} & \parbox{4.5cm}{\texttt{nvidia/Nemotron-\\Research-Reasoning-Qwen-1.5B}} & ~\citep{liu2025prorl} \\ \\
OpenT & OpenThinker-7B & \parbox{4.5cm}{\texttt{open-thoughts/ \\ OpenThinker-7B}} & ~\citep{guha2025openthoughts} \\ \\
OSS & GPT-OSS-20B & \parbox{4.5cm}{\texttt{openai/gpt-oss-20b}} & ~\citep{openai2025gpt} \\
 \\
QwQ & Qwen/QwQ-32B & \parbox{4.5cm}{\texttt{Qwen/QwQ-32B}} & ~\citep{qwq32b} \\ \\
DAPO & DAPO-Qwen-32B & \parbox{4.5cm}{\texttt{BytedTsinghua-SIA/\\DAPO-Qwen-32B}} & ~\citep{yu2025dapoopensourcellmreinforcement} \\
\bottomrule
\end{tabular}
}
\label{tab:model_info}
\end{table}

\begin{table}[t]
\centering
\caption{Model configurations and reasoning approaches evaluated in the user study.}
\label{tab:user_study_model_info}
\small
\begin{tabular}{lp{8cm}}
\toprule
\textbf{Reasoning Approach} & \textbf{Model Configuration} \\
\midrule
Default CoT (Greedy) & GPT-OSS-20B \\
Default CoT (Greedy) & DAPO-Qwen-32B \\
Ensemble CoT (Generator / Evaluator) & QwQ-32B + DAPO-Qwen-32B / GPT-OSS-20B \\
Ensemble CoT (Generator / Evaluator) & QwQ-32B + GPT-OSS-20B / DAPO-Qwen-32B \\
\bottomrule
\end{tabular}
\end{table}

\section{Results}

\subsection{Evaluating generalizability of LRM CoT explanations}

\begin{figure}[!t]
    \centering
    \includegraphics[width=\linewidth]{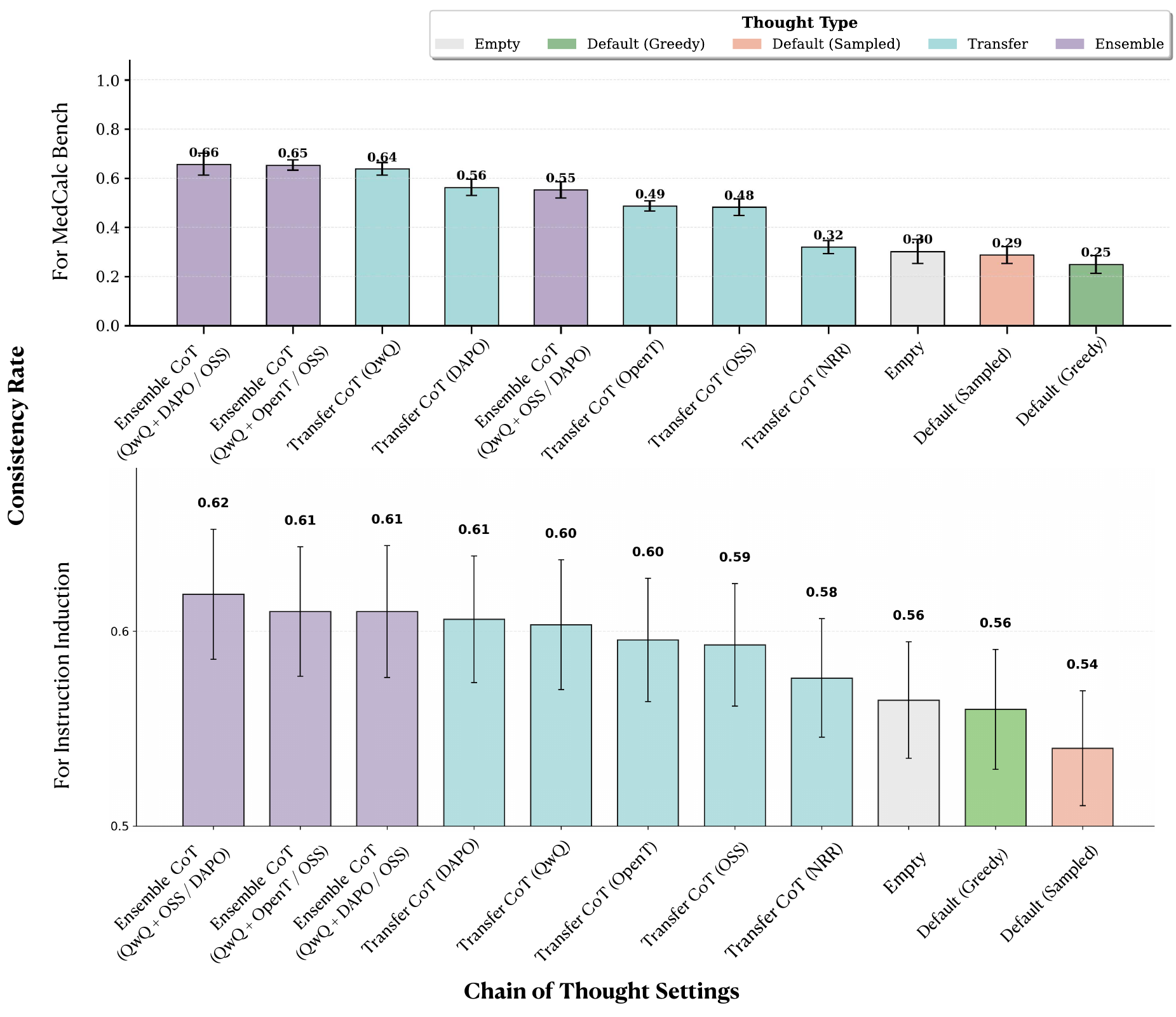}
    \caption{Average pairwise consistency across thought settings in MedCalc-Bench (above) and Instruction Induction (below). For thought variations indicating Ensemble CoT, models listed before the slash (/) serve as generators, while the model after the slash acts as the judge/evaluator.
    }
    \label{fig:consistency-barplot}
\end{figure}

\begin{tcolorbox}[colback=cyan!10, colframe=black, boxrule=1pt, title=Finding 1]
LRM explanations generalize to other LRMs, even when the explanation induces an inaccurate answer.
\end{tcolorbox}
\paragraph{Cross-model consistency} \cref{fig:consistency-barplot} shows that providing LRMs with shared CoT explanations dramatically increases their consistency, i.e. how often they arrive at the same answer.
Transfer and ensemble CoTs substantially improve consistency in both benchmarks. On MedCalc-Bench, consistency increases from a 25\% baseline to 66\%. Instruction-Induction shows similar gains, rising from 54\% to 62\% with ensemble CoT.

Interestingly, the default CoT with sampling outperforms greedy default CoT consistency in the MedCalc-Bench setting, which suggests that stochastic regularization may aid cross-model alignment for this calculation-heavy task. However, this advantage is not present in the Instruction-Induction setting, where sampled and greedy achieve comparable consistency rates. This suggests that the effectiveness of different decoding strategies for promoting cross-model consistency may depend on task characteristics.

\begin{figure}[!t]
    \centering
    \includegraphics[width=\linewidth]{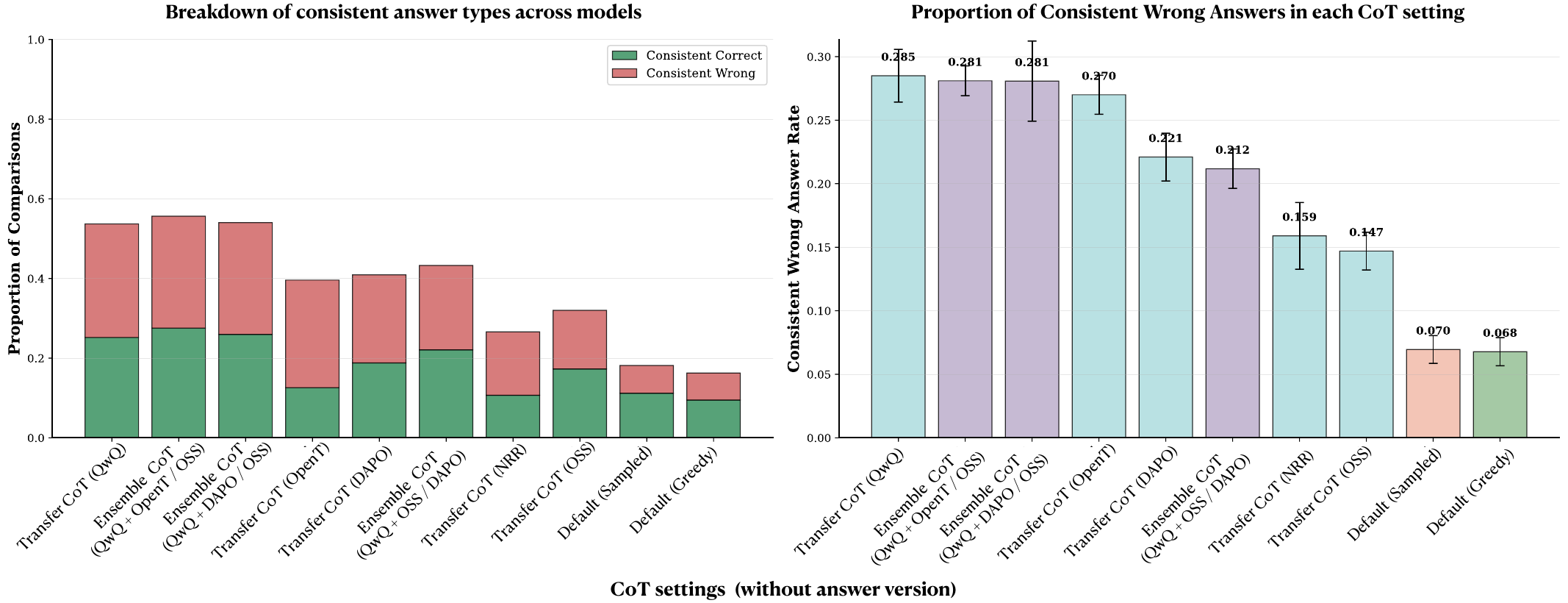}
    \caption{\textbf{Consistency breakdown across thought variations} \textit{Left:} Proportion of consistent outputs separated into matching correct and matching incorrect conclusions.  \textit{Right:} Rate of consistent answers that are wrong across various thought settings.
    }
    \label{fig:wrong-answer-analysis}
\end{figure}

The increase in consistency includes matching `incorrect' answers, where models arrive at the same wrong conclusion when provided with identical chains-of-thought.
\cref{fig:wrong-answer-analysis} shows how often a CoT leads models to converge on the same answer even when the predicted answer is not mentioned in the CoT (i.e., when the CoT mostly provides partial hints or reasoning) in the MedCalc-Bench setting. The left panel breaks down these “same answer” cases into convergence on the same correct versus the same incorrect answers. The right panel shows the proportion of consistently incorrect answers across CoT types. These results demonstrate that CoTs can systematically steer model reasoning, including toward the same conclusion, which indicates that CoT can exert a generalizing influence on model behavior even when the reasoning they provide can be incorrect. 

\begin{table}[!t]
\centering
\footnotesize
\caption{Comparison of CoT accuracy across models for MedCalc-Bench and Instruction Induction.
See \cref{sec:remove_answer_method} and \cref{sec:additional_analysis} for more details.}
\begin{center}
\label{tab:accuracy}
\scriptsize
\setlength{\tabcolsep}{3pt}
\begin{tabular}{@{}p{3cm} *{5}{c}  c | *{5}{c} c}
\toprule
\multirow{3}{*}{\textbf{Method}} & \multicolumn{5}{c}{\textbf{MedCalc-Bench (Exact-Match)}} & \multirow{3}{*}{\textbf{Avg}} & \multicolumn{5}{c}{\textbf{Instruction Induction (BERTScore)}} & \multirow{3}{*}{\textbf{Avg}} \\
\cmidrule(lr){2-6} \cmidrule(lr){8-12}
 & \textbf{NRR} & \textbf{OpenT} & \textbf{OSS} & \textbf{QwQ} & \textbf{DAPO} & & \textbf{NRR} & \textbf{OpenT} & \textbf{OSS} & \textbf{QwQ} & \textbf{DAPO} & \\
 & \textbf{1.5B} & \textbf{7B} & \textbf{20B} & \textbf{32B} & \textbf{32B} & & \textbf{1.5B} & \textbf{7B} & \textbf{20B} & \textbf{32B} & \textbf{32B} & \\
\midrule
 Empty CoT & 0.10 & 0.18 & 0.45 & 0.36 & 0.38 & 0.29 & 0.53 & 0.55 & 0.56 & 0.55 & 0.57 & 0.55 \\
\midrule
\multirow{1}{*}{Default (Greedy) CoT} 
  & 0.14 & 0.24 & 0.43 & 0.38 & 0.41 & 0.32 & 0.58 & 0.46 & 0.61 & 0.60 & 0.62 & 0.57 \\
\midrule
\multirow{1}{*}{Default (Sampled) CoT} 
  & 0.16 & 0.29 & 0.45 & 0.38 & 0.35 & 0.33 & 0.56 & 0.56 & 0.60 & 0.60 & 0.60 & 0.58 \\
\midrule
\multirow{1}{*}{Trans. CoT (NRR)} 
   & 0.14 & 0.15 & 0.24 & 0.30 & 0.34 & 0.23 & 0.58 & 0.57 & 0.60 & 0.60 & 0.63 & 0.60 \\
\midrule
\multirow{1}{*}{Trans. CoT (OpenT)} 
   & 0.21 & 0.24 & 0.26 & 0.26 & 0.25 & 0.24 & 0.60 & 0.46 & 0.59 & 0.59 & 0.61 & 0.57 \\
\midrule
\multirow{1}{*}{Trans. CoT (OSS)} 
   & 0.26 & 0.44 & 0.43 & 0.40 & 0.44 & 0.39 & 0.60 & 0.57 & 0.61 & 0.60 & 0.62 & 0.60 \\
\midrule
\multirow{1}{*}{Trans. CoT (QwQ)} 
   & 0.34 & 0.37 & 0.39 & 0.38 & 0.37 & 0.37 & 0.60 & 0.57 & 0.61 & 0.60 & 0.62 & 0.60 \\
\midrule
\multirow{1}{*}{Trans. CoT (DAPO)} 
   & 0.31 & 0.40 & 0.39 & 0.40 & 0.41 & 0.38 & 0.60 & 0.58 & 0.62 & 0.61 & 0.62 & 0.61 \\
\midrule
\multirow{1}{*}{Ens. (QwQ+DAPO/OSS)} 
   & 0.39 & 0.37 & 0.41 & 0.41 & 0.40 & 0.40 & 0.60 & 0.57 & 0.61 & 0.60 & 0.62 & 0.60 \\
\midrule
\multirow{1}{*}{Ens. (QwQ+OSS/DAPO)} 
   & 0.28 & 0.37 & 0.45 & 0.42 & 0.43 & 0.38 & 0.60 & 0.57 & 0.61 & 0.60 & 0.62 & 0.60 \\
\midrule
\multirow{1}{*}{Ens. (QwQ+OpenT/OSS)} 
  & 0.34 & 0.41 & 0.42 & 0.38 & 0.40 & 0.39 & 0.61 & 0.58 & 0.61 & 0.60 & 0.62 & 0.60 \\
\bottomrule
\end{tabular}
\end{center}

\end{table}

\paragraph{Which LRMs work best when extracting CoTs from an ensemble?}
The strongest performance comes from specific combinations of ensemble methods and thought variations in both MedCalc-Bench and Instruction Induction. In MedCalc-Bench, ensembles that use OSS as the evaluator achieve substantially higher consistency than other configurations. When OSS serves as the generator, consistency decreases, remaining above OSS on its own but closer to the lower end of the distribution. In Instruction-Induction, transferring OSS's CoT yields the strongest performance compared to other transfers. Similarly, the ensemble that uses OSS as the generator outperforms the other ensemble configurations. Taken together, these results suggest that models whose CoT transfers exhibit greater consistency also tend to function as more effective generators within ensemble transfers.

\begin{figure}[!ht]
    \centering
    \includegraphics[width=\linewidth]{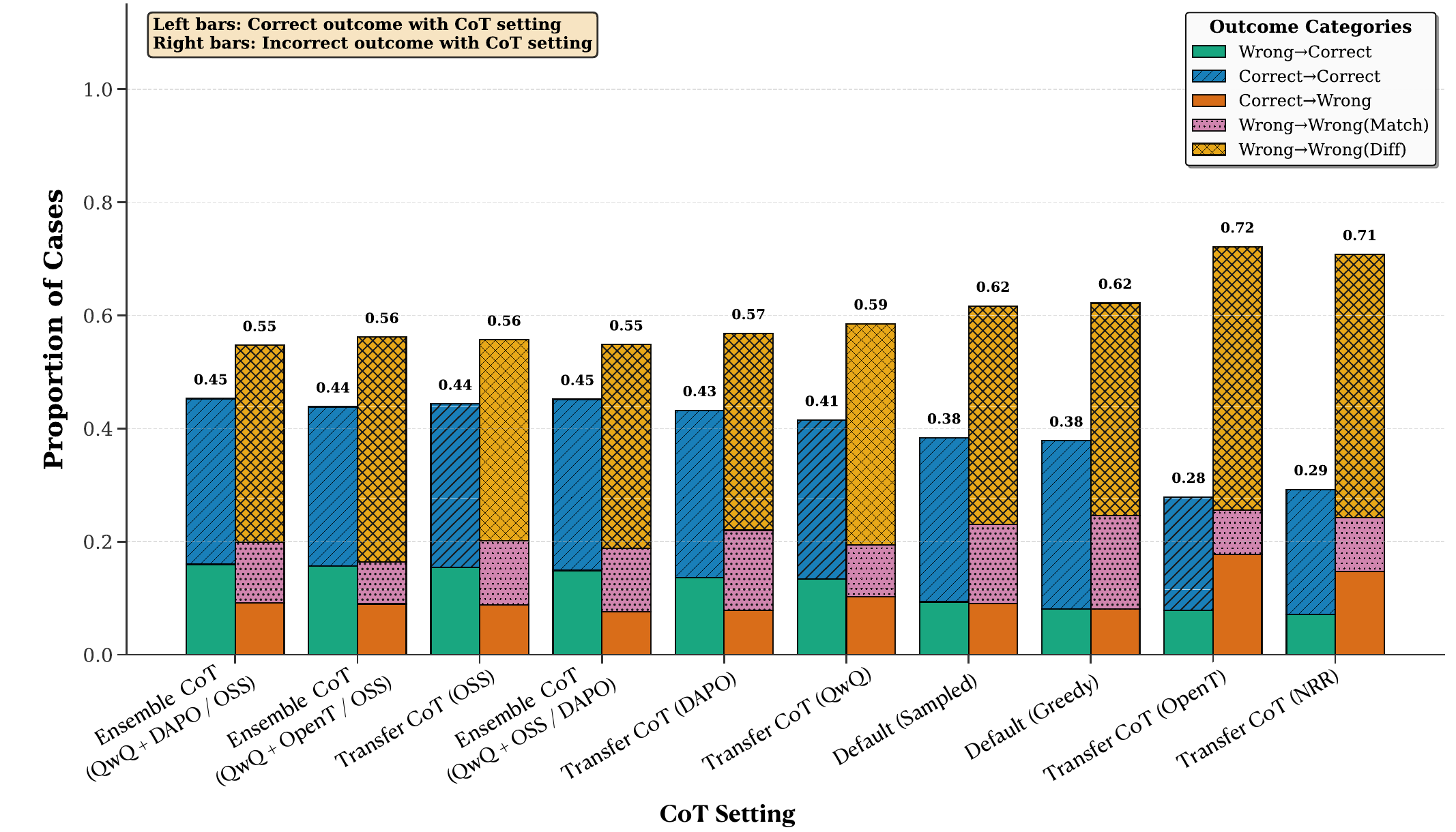}
    \caption{\textbf{CoT Transfer Effect Analysis} Distribution of transfer outcomes when CoT reasoning is used or transferred across models. Each CoT setting  is evaluated by comparing model predictions with CoT
    versus without CoT (empty baseline). Settings include, Default: models using their own generated CoT; Sampled: CoT generated through sampling; Transfer CoT $l_{gen}$: CoT transferred from model $l_{gen}$ to all models; Ensemble CoT: combined CoT from multiple models. The five conditions represent: Wrong→Correct: cases where CoT successfully corrects errors (green); Correct→Wrong: cases where CoT misleads the model from correct to incorrect predictions (red); Correct→Correct: cases where CoT maintains correct predictions (blue); Wrong→Wrong(Match): both predictions incorrect with identical wrong answers (light gray); Wrong→Wrong(Diff): both predictions incorrect with different wrong answers (dark gray). Results are aggregated across multiple target models for each CoT setting. Settings are sorted by Wrong→Correct rate (descending).}
    \label{fig:ablation-analysis}
\end{figure}

\paragraph{Within-model prediction shifts across CoT conditions} \cref{fig:ablation-analysis} shows a breakdown of outcomes comparing CoT and baseline empty CoT reasoning across several scenarios, including cases where (a) an incorrect model prediction becomes correct after CoT transfer, (b) a correct prediction is perturbed, and (c) different forms of agreement or disagreement emerge across models. Regarding accuracy, \cref{tab:accuracy} shows the effect of these chains-of-thought on different models. The takeaway is that models with weaker baseline accuracy can reduce the accuracy of other models when their CoTs are transferred even if the other model inherently performs better at their own baseline.

\begin{figure}[!ht]
    \centering
    \includegraphics[width=\linewidth]{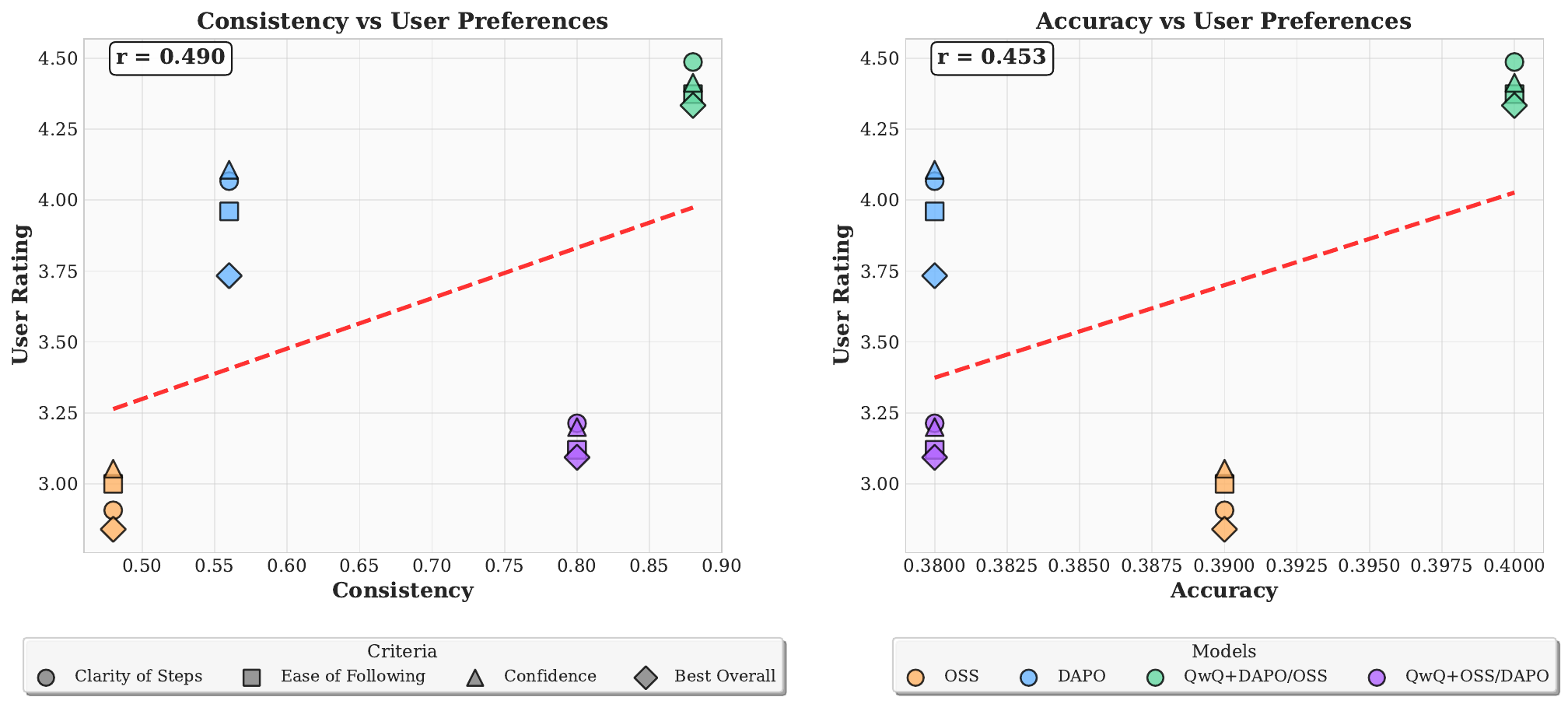}
   \caption{Comparing human user ratings of CoTs to LRM consistency (left) and LRM accuracy (right).
   Consistency and accuracy both show positive correlations with user ratings, with consistency showing a stronger trend.
   User ratings were collected for four criteria: Clarity of Steps, Ease of Following, Confidence, and Best Overall.
   }
    \label{fig:user-study-plot}
\end{figure}

\subsection{User Study}

\begin{tcolorbox}[colback=cyan!10, colframe=black, boxrule=1pt, title=Finding 2]
LRM explanations that generalize to other LRMs receive higher user preference ratings.
\end{tcolorbox}

The user study results suggest that improving the two metrics of CoT generalizability can enhance perceived model explanation quality.
(see \cref{fig:user-study-plot}), although they should be interpreted with caution as they include only 4 LRM combinations.
Of the two metrics, consistency appears to be a stronger predictor of user satisfaction than accuracy.

The user study results further show that the default DAPO CoT and the ensembled QwQ+DAPO/OSS CoT were consistently perceived as easier to understand than the default OSS CoT and the ensembled QwQ+OSS/DAPO CoT.
Independent t-tests with Bonferroni correction confirmed that OSS was rated significantly worse than both DAPO ($p < 0.0001$) and QwQ+DAPO/OSS ($p < 0.0001$) in terms of \textit{Clarity of Steps}. This pattern extended to \textit{Ease of Following} and \textit{Confidence}. In contrast, one ensemble variant performed similarly to OSS, and the difference was not statistically significant ($p = 1.0$). All other comparison results, including those for \textit{Best Overall}, showcased significant differences.
Between DAPO and ensembled QwQ+DAPO/OSS CoT, no significant differences emerged for \textit{Clarity of Steps}, \textit{Ease of Following} and \textit{Confidence}.
For \textit{Best Overall}, the ensemble was rated significantly higher ($p = 0.005$), suggesting that the ensemble with DAPO as generator and OSS as evaluator is the most effective configuration.
See a further breakdown of results in \cref{fig:user-study-plot-appendix}.

\subsection{Implications for RL Training}
\begin{tcolorbox}[colback=cyan!10, colframe=black, boxrule=1pt, title=Finding 3]
RL post-training that improves an LRM's performance improves its CoT consistency, but not necessarily its CoT transfer accuracy.
\end{tcolorbox}
To assess the effect of reinforcement learning on model consistency and accuracy, we post-train two base models (Deepseek-R1-Distill-Qwen-1.5B and Llama-3.2-3B-Instruct) on the MedCalc-Bench dataset.
We follow the RLVR-only setup and hyperparameters described for this dataset in~\citep{lin2025trainingllmsehrbasedreasoning}.

\begin{table}[!ht]
\centering
\caption{Effect of GRPO training on consistency and accuracy. Self-accuracy measures each model's own performance. Avg. accuracy reports the mean accuracy when transferring each LRM's CoT to five other models (NRR, OpenT, OSS, QwQ, DAPO).
Avg. consistency measures mean pairwise consistency across the five models when evaluating each model's transferred reasoning.}
\label{tab:grpo_comparison}
\begin{tabular}{@{}lccc@{}}
\toprule
\textbf{Model} & \textbf{Self-Accuracy} & \makecell{\textbf{Avg. Accuracy}\\(Transfer CoT)} & \makecell{\textbf{Avg. Consistency}\\(Transfer CoT)} \\
\midrule
Deepseek-R1-Distill-Qwen-1.5B (Base) & 0.11 $\pm$ 0.03 & 0.30 $\pm$ 0.052 & 0.11 $\pm$ 0.007  \\
Deepseek-R1-Distill-Qwen-1.5B (GRPO) & 0.18 $\pm$ 0.04 & 0.29 $\pm$ 0.044  & 0.39  $\pm$ 0.061  \\
\midrule
Llama-3.2-3B-Instruct (Base) & 0.12 $\pm$ 0.03  & 0.25 $\pm$ 0.050 & 0.25 $\pm$ 0.019 \\
Llama-3.2-3B-Instruct (GRPO) & 0.45 $\pm$ 0.05 & 0.44 $\pm$ 0.039 & 0.46 $\pm$ 0.035 \\
\bottomrule
\end{tabular}
\end{table}

\cref{tab:grpo_comparison} reveals a notable difference between consistency and accuracy improvements, after RL post-training. 
While both models show substantial consistency gains, their accuracy trajectories diverge: Deepseek's average accuracy remains flat despite improved self-accuracy, while Llama shows coupled improvements across all metrics.
This demonstrates that consistency and accuracy are separable properties that may vary independently.
Future training methods may benefit from explicitly optimizing both consistency and transfer accuracy to enhance the generalizability of chain-of-thought reasoning across models.

\section{Related work}
\paragraph{Generating and improving natural-language explanations}
A large body of work extends CoT prompting~\citep{wei2022chain} by probing or refining the explanations it produces.
Examples include evaluating counterfactuals introduced into the CoT~\citep{gat2023faithful}, testing their robustness to mistakes introduced into the reasoning chain~\citep{lanham2023measuring},
or using contrastive CoT to induce reliance on the reasoning chain~\citep{chia2023contrastive}.
A few works seek to improve the consistency in the generations made by an LLM, either between the generation and validation of LLMs~\citep{li2023benchmarking},
between LLM predictions on implications of an original question~\citep{akyurek2024deductive},
on counterfactual inputs for an original question~\citep{chen2025towards,shihab2025counterfactual},
or by more generally introducing desirable structures into reasoning traces~\citep{sun2025relif}.

A similar line of work has studied generating explanations directly for a problem/dataset, rather than for a single example, e.g. describing distributions in natural language~\citep{zhong2023goaldd,singh2023explaining} or human-readable programs~\citep{romera2024mathematical,novikov2025alphaevolve}.
These works rely on some form of external verification for explanations (e.g. restricting an explanation to be python-runnable code) rather than allowing them to be flexible. A separate line of work has studied ensemble LLM generation~\citep{tekin2024llm,chen2025harnessing}, although not at the sentence-level and not for the purpose of explanation generation.

\paragraph{Assessing CoT explanations}
Model-generated text explanations have shown issues with faithfulness to the underlying LLM/LRM~\citep{turpin2023language,ye-durrett-2022-explanations}, e.g. LLM reasoning chains have been shown to be inconsistent across counterfactuals~\citep{mancoridis2025potemkin},
sensitive to minor variations~\citep{yeo2024interpretable},
the answer may not follow from the chain~\citep{xiong2025measuring},
may not reveal the info they really rely on~\citep{chen2025reasoning},
inconsistently learn algorithms~\citep{shojaee2025illusion},
can succeed at reasoning with invalid intermediate tokens~\citep{stechly2025beyond},
or be trained to use dummy intermediate tokens~\citep{pfau2024lets}.
Additionally, human studies suggest that users perceive the wrong narratives from reasoning chains \citep{levy2025humans} and that users do not necessarily rank accurate reasoning traces for models higher~\citep{bhambri2025cognitively}.
See also other warnings about relying on LLM reasoning traces~\citep{kambhampati2025stop,bhambri2025interpretable,chua2025deepseek}, including mechanistic analysis~\citep{bogdan2025thought,prakash2025language}, and on the difficulty of evaluating reasoning faithfulness~\citep{zaman2025causal} and monitoring chain-of-thoughts~\citep{korbak2025chainthoughtmonitorabilitynew, guan2025monitoringmonitorability}.

\paragraph{Evaluating natural-language explanations}
Prior works for evaluating natural-language explanations have aligned on one of three dimensions: consistency, plausibility, and faithfulness. 
\textit{Consistency}, which we focus on in this work, measures if the model generates consistent explanations on similar examples \citep{hase-bansal-2020-evaluating, chen2023models}.
\textit{Plausibility} evaluates humans' preference of an explanation based on its factual correctness and logical coherence \citep{herman2017promise, lage2019evaluation, jacovi-goldberg-2020-towards}.
It is different from \textit{faithfulness}, which measures whether an explanation is consistent with the model's internal decision process \citep{harrington1985harvey,
10.1145/2939672.2939778,
gilpin2018explaining,
jacovi-goldberg-2020-towards}.
 More broadly, explanation evaluation frameworks such as ~\citep{doshi2017towards}, ~\citep{10.1145/2939672.2939778}, and surveys such as~\citep{zhou2021evaluating, hoffman2019metricsexplainableaichallenges} emphasize the distinction between human-centered and model-centered explanation quality and the need for metrics that reflect different goals of interpretability. We extend these metrics by evaluating explanations across models rather than within a single model. We measure whether a CoT from one LRM generalizes behaviorally to others through cross-model consistency. This provides a complementary perspective to existing explanation-evaluation frameworks focused on human preference or model faithfulness.

\FloatBarrier
\section{Discussion}

Our analysis is motivated by the question, \textit{``What is a good explanation?''}
Unlike previous work that has focused on faithfulness or correctness, we focus on the question of generalization: the notion that a good explanation should be effective at guiding a new user as the explainer intended.
Taking advantage of the structure of LRMs as both producers and users of CoT, we establish a framework for approaching this problem by measuring the generalization of CoT from one LRM to another.

While the focus here is on cross-LRM generalization,
we note that this measure has some bearing on whether explanations generalize to humans (as seen in our human survey).
As LRMs become increasingly adept at simulating a user~\citep{dou2025simulatorarena,naous2025flipping},
the same framework can help us evaluate an explanation's generalizability to humans.
However, currently this assumption may fail in the case that different LRMs share a common bias for a particular explanation.

The framework here lays the groundwork for a variety of future work.
One important domain in AI safety may seek generalizable CoTs for the purpose of monitoring and understanding by LLM-based safety tools~\citep{bedi2025fidelity,zhao2025chain}.
Another line of work may seek to find CoTs that generalize across diverse examples for the purpose of generating new knowledge~\citep{venkatraman2025recursive,qu2025rlad,feng2026hachi}.
Finally,
future work could explore improvements in explanation generation with ensembles (building on the straightforward strategy introduced here) or in improving RL post-training pipelines to incentivize generating generalizable explanations.
We hope that properly generating and evaluating generalizable explanations produced by LRM can ultimately help humans understand subjects beyond current human knowledge.

\section*{Acknowledgments}
The authors thank Chris Ackerman, Lace Padilla, and Byron Wallace for advice on this project, as well as the members of the Baulab and Millicent Li for their feedback on the paper. We also thank the participants in our user study, whose contributions provided valuable insights for this work. This research was supported by the generous funding of the Cambridge-Boston Alignment Initiative (CBAI) Fellowship and Coefficient Giving.

\section*{Reproducibility}
\label{sec:repro}

All experiments were run either on workstations with
141GB NVIDIA H200 SXM GPUs or 80GB NVIDIA A100 GPUs using the HuggingFace Transformers library \citep{wolf-etal-2020-transformers}. The code and dataset produced during this work is publicly available here: \url{https://genex.baulab.info}.

\section*{Ethics}
\label{sec:ethics}

Our work studies the generalizability of chain-of-thought (CoT) reasoning across different models and tasks. While CoT can improve performance and interpretability, its generalizability should be considered carefully. Reasoning patterns that transfer well in one setting may also reinforce shared mistakes in another, leading to consistent but incorrect outputs. In addition, reusing or combining CoTs across models may affect accuracy in ways that are not always predictable. These effects are particularly important to keep in mind in sensitive application areas, such as healthcare or law, where errors carry higher risks. We view this study as a step toward understanding both the benefits and limitations of CoT transfer. Future work should continue to explore when and how CoT generalizes reliably, and how to identify cases where it may not. In the user study we conducted, no personal information was collected during the user study experiments.

\section*{Use of Large Language Models}
\label{sec:use-of-llms}

We used LLMs to help with coding for plots and minor editing of paper text.

{
    \small
    \bibliography{refs.bib}

\begin{thebibliography}{72}
\providecommand{\natexlab}[1]{#1}
\providecommand{\url}[1]{\texttt{#1}}
\expandafter\ifx\csname urlstyle\endcsname\relax
  \providecommand{\doi}[1]{doi: #1}\else
  \providecommand{\doi}{doi: \begingroup \urlstyle{rm}\Url}\fi

\bibitem[Abdin et~al.(2025)Abdin, Agarwal, Awadallah, Balachandran, Behl, Chen,
  de~Rosa, Gunasekar, Javaheripi, Joshi, et~al.]{abdin2025phi}
Marah Abdin, Sahaj Agarwal, Ahmed Awadallah, Vidhisha Balachandran, Harkirat
  Behl, Lingjiao Chen, Gustavo de~Rosa, Suriya Gunasekar, Mojan Javaheripi,
  Neel Joshi, et~al.
\newblock Phi-4-reasoning technical report.
\newblock \emph{arXiv preprint arXiv:2504.21318}, 2025.

\bibitem[Agarwal et~al.(2025)Agarwal, Ahmad, Ai, Altman, Applebaum, Arbus,
  Arora, Bai, Baker, Bao, et~al.]{agarwal2025gpt}
Sandhini Agarwal, Lama Ahmad, Jason Ai, Sam Altman, Andy Applebaum, Edwin
  Arbus, Rahul~K Arora, Yu~Bai, Bowen Baker, Haiming Bao, et~al.
\newblock gpt-oss-120b \& gpt-oss-20b model card.
\newblock \emph{arXiv preprint arXiv:2508.10925}, 2025.

\bibitem[Aky{\"u}rek et~al.(2024)Aky{\"u}rek, Aky{\"u}rek, Choshen, Wijaya, and
  Andreas]{akyurek2024deductive}
Afra~Feyza Aky{\"u}rek, Ekin Aky{\"u}rek, Leshem Choshen, Derry Wijaya, and
  Jacob Andreas.
\newblock Deductive closure training of language models for coherence,
  accuracy, and updatability.
\newblock \emph{arXiv preprint arXiv:2401.08574}, 2024.

\bibitem[Barez et~al.(2025)Barez, Wu, Arcuschin, Lan, Wang, Siegel, Collignon,
  Neo, Lee, Paren, et~al.]{barez2025chain}
Fazl Barez, Tung-Yu Wu, Iv{\'a}n Arcuschin, Michael Lan, Vincent Wang, Noah
  Siegel, Nicolas Collignon, Clement Neo, Isabelle Lee, Alasdair Paren, et~al.
\newblock Chain-of-thought is not explainability.
\newblock \emph{Preprint, alphaXiv}, pp.\ ~v2, 2025.

\bibitem[Bedi et~al.(2025)Bedi, Jiang, Chung, Koyejo, and
  Shah]{bedi2025fidelity}
Suhana Bedi, Yixing Jiang, Philip Chung, Sanmi Koyejo, and Nigam Shah.
\newblock Fidelity of medical reasoning in large language models.
\newblock \emph{JAMA Network Open}, 8\penalty0 (8):\penalty0
  e2526021--e2526021, 2025.

\bibitem[Bewersdorff et~al.(2025)Bewersdorff, Hartmann, Hornberger, Se{\ss}ler,
  Bannert, Kasneci, Kasneci, Zhai, and Nerdel]{bewersdorff2025taking}
Arne Bewersdorff, Christian Hartmann, Marie Hornberger, Kathrin Se{\ss}ler,
  Maria Bannert, Enkelejda Kasneci, Gjergji Kasneci, Xiaoming Zhai, and Claudia
  Nerdel.
\newblock Taking the next step with generative artificial intelligence: The
  transformative role of multimodal large language models in science education.
\newblock \emph{Learning and Individual Differences}, 118:\penalty0 102601,
  2025.

\bibitem[Bhambri et~al.(2025{\natexlab{a}})Bhambri, Biswas, and
  Kambhampati]{bhambri2025cognitively}
Siddhant Bhambri, Upasana Biswas, and Subbarao Kambhampati.
\newblock Do cognitively interpretable reasoning traces improve llm
  performance?
\newblock \emph{arXiv preprint arXiv:2508.16695}, 2025{\natexlab{a}}.

\bibitem[Bhambri et~al.(2025{\natexlab{b}})Bhambri, Biswas, and
  Kambhampati]{bhambri2025interpretable}
Siddhant Bhambri, Upasana Biswas, and Subbarao Kambhampati.
\newblock Interpretable traces, unexpected outcomes: Investigating the
  disconnect in trace-based knowledge distillation, 2025{\natexlab{b}}.
\newblock URL \url{https://arxiv.org/abs/2505.13792}.

\bibitem[Bogdan et~al.(2025)Bogdan, Macar, Nanda, and Conmy]{bogdan2025thought}
Paul~C Bogdan, Uzay Macar, Neel Nanda, and Arthur Conmy.
\newblock Thought anchors: Which llm reasoning steps matter?
\newblock \emph{arXiv preprint arXiv:2506.19143}, 2025.

\bibitem[Chen et~al.(2024)Chen, Zhong, Ri, Zhao, He, Steinhardt, Yu, and
  McKeown]{chen2023models}
Yanda Chen, Ruiqi Zhong, Narutatsu Ri, Chen Zhao, He~He, Jacob Steinhardt, Zhou
  Yu, and Kathleen McKeown.
\newblock Do models explain themselves? counterfactual simulatability of
  natural language explanations.
\newblock In \emph{Proceedings of the 41st International Conference on Machine
  Learning}, pp.\  7880--7904, 2024.

\bibitem[Chen et~al.(2025{\natexlab{a}})Chen, Benton, Radhakrishnan, Uesato,
  Denison, Schulman, Somani, Hase, Wagner, Roger, et~al.]{chen2025reasoning}
Yanda Chen, Joe Benton, Ansh Radhakrishnan, Jonathan Uesato, Carson Denison,
  John Schulman, Arushi Somani, Peter Hase, Misha Wagner, Fabien Roger, et~al.
\newblock Reasoning models don't always say what they think.
\newblock \emph{arXiv preprint arXiv:2505.05410}, 2025{\natexlab{a}}.

\bibitem[Chen et~al.(2025{\natexlab{b}})Chen, Singh, Liu, Zuo, Yu, He, and
  Gao]{chen2025towards}
Yanda Chen, Chandan Singh, Xiaodong Liu, Simiao Zuo, Bin Yu, He~He, and
  Jianfeng Gao.
\newblock Towards consistent natural-language explanations via
  explanation-consistency finetuning.
\newblock In \emph{Proceedings of the 31st International Conference on
  Computational Linguistics}, pp.\  7558--7568, 2025{\natexlab{b}}.

\bibitem[Chen et~al.(2025{\natexlab{c}})Chen, Li, Chen, Li, Sun, Luo, Mao,
  Yang, Sun, and Yu]{chen2025harnessing}
Zhijun Chen, Jingzheng Li, Pengpeng Chen, Zhuoran Li, Kai Sun, Yuankai Luo,
  Qianren Mao, Dingqi Yang, Hailong Sun, and Philip~S Yu.
\newblock Harnessing multiple large language models: A survey on llm ensemble.
\newblock \emph{arXiv preprint arXiv:2502.18036}, 2025{\natexlab{c}}.

\bibitem[Chia et~al.(2023)Chia, Chen, Tuan, Poria, and
  Bing]{chia2023contrastive}
Yew~Ken Chia, Guizhen Chen, Luu~Anh Tuan, Soujanya Poria, and Lidong Bing.
\newblock Contrastive chain-of-thought prompting, 2023.

\bibitem[Chua \& Evans(2025)Chua and Evans]{chua2025deepseek}
James Chua and Owain Evans.
\newblock Are deepseek r1 and other reasoning models more faithful?
\newblock \emph{arXiv preprint arXiv:2501.08156}, 2025.

\bibitem[Doshi-Velez \& Kim(2017)Doshi-Velez and Kim]{doshi2017towards}
Finale Doshi-Velez and Been Kim.
\newblock Towards a rigorous science of interpretable machine learning.
\newblock \emph{ArXiv}, 2017.
\newblock URL \url{https://arxiv.org/pdf/1702.08608.pdf}.

\bibitem[Dou et~al.(2025)Dou, Galley, Peng, Kedzie, Cai, Ritter, Quirk, Xu, and
  Gao]{dou2025simulatorarena}
Yao Dou, Michel Galley, Baolin Peng, Chris Kedzie, Weixin Cai, Alan Ritter,
  Chris Quirk, Wei Xu, and Jianfeng Gao.
\newblock Simulatorarena: Are user simulators reliable proxies for multi-turn
  evaluation of ai assistants?
\newblock In \emph{Proceedings of the 2025 Conference on Empirical Methods in
  Natural Language Processing}, pp.\  35200--35278, 2025.

\bibitem[Feng et~al.(2026)Feng, Kothari, Vossler, Bishara, Zier, Addo,
  Kornblith, Tan, and Singh]{feng2026hachi}
Jean Feng, Avni Kothari, Patrick Vossler, Andrew Bishara, Lucas Zier, Newton
  Addo, Aaron Kornblith, Yan~Shuo Tan, and Chandan Singh.
\newblock Human-ai co-design for clinical prediction models, 2026.
\newblock URL \url{https://arxiv.org/abs/2601.09072}.

\bibitem[Gat et~al.(2023)Gat, Calderon, Feder, Chapanin, Sharma, and
  Reichart]{gat2023faithful}
Yair Gat, Nitay Calderon, Amir Feder, Alexander Chapanin, Amit Sharma, and Roi
  Reichart.
\newblock Faithful explanations of black-box nlp models using llm-generated
  counterfactuals.
\newblock \emph{arXiv preprint arXiv:2310.00603}, 2023.

\bibitem[Gilpin et~al.(2018)Gilpin, Bau, Yuan, Bajwa, Specter, and
  Kagal]{gilpin2018explaining}
Leilani~H Gilpin, David Bau, Ben~Z Yuan, Ayesha Bajwa, Michael Specter, and
  Lalana Kagal.
\newblock Explaining explanations: An approach to evaluating interpretability
  of machine learning.
\newblock \emph{arXiv preprint arXiv:1806.00069}, 2018.

\bibitem[Guan et~al.(2025)Guan, Wang, Carroll, Dou, Wei, Williams, Arnav,
  Huizinga, Kivlichan, Glaese, Pachocki, and
  Baker]{guan2025monitoringmonitorability}
Melody~Y. Guan, Miles Wang, Micah Carroll, Zehao Dou, Annie~Y. Wei, Marcus
  Williams, Benjamin Arnav, Joost Huizinga, Ian Kivlichan, Mia Glaese, Jakub
  Pachocki, and Bowen Baker.
\newblock Monitoring monitorability, 2025.
\newblock URL \url{https://arxiv.org/abs/2512.18311}.

\bibitem[Guha et~al.(2025)Guha, Marten, Keh, Raoof, Smyrnis, Bansal, Nezhurina,
  Mercat, Vu, Sprague, et~al.]{guha2025openthoughts}
Etash Guha, Ryan Marten, Sedrick Keh, Negin Raoof, Georgios Smyrnis, Hritik
  Bansal, Marianna Nezhurina, Jean Mercat, Trung Vu, Zayne Sprague, et~al.
\newblock Openthoughts: Data recipes for reasoning models.
\newblock \emph{arXiv preprint arXiv:2506.04178}, 2025.

\bibitem[Guo et~al.(2025)Guo, Yang, Zhang, Song, Zhang, Xu, Zhu, Ma, Wang, Bi,
  et~al.]{guo2025deepseek}
Daya Guo, Dejian Yang, Haowei Zhang, Junxiao Song, Ruoyu Zhang, Runxin Xu,
  Qihao Zhu, Shirong Ma, Peiyi Wang, Xiao Bi, et~al.
\newblock Deepseek-r1: Incentivizing reasoning capability in llms via
  reinforcement learning.
\newblock \emph{arXiv preprint arXiv:2501.12948}, 2025.

\bibitem[Harrington et~al.(1985)Harrington, Morley, {\v{S}}cedrov, and
  Simpson]{harrington1985harvey}
Leo~A Harrington, Michael~D Morley, A~{\v{S}}cedrov, and Stephen~G Simpson.
\newblock \emph{Harvey Friedman's research on the foundations of mathematics}.
\newblock 1985.
\newblock URL
  \url{https://books.google.com/books/about/Harvey_Friedman_s_Research_on_the_Founda.html?id=2plPRR4LDxIC}.

\bibitem[Hase \& Bansal(2020)Hase and Bansal]{hase-bansal-2020-evaluating}
Peter Hase and Mohit Bansal.
\newblock Evaluating explainable {AI}: Which algorithmic explanations help
  users predict model behavior?
\newblock In \emph{Proceedings of the Association for Computational
  Linguistics}, 2020.
\newblock URL \url{https://aclanthology.org/2020.acl-main.491}.

\bibitem[Herman(2017)]{herman2017promise}
Bernease Herman.
\newblock The promise and peril of human evaluation for model interpretability.
\newblock \emph{ArXiv}, 2017.
\newblock URL \url{https://arxiv.org/pdf/1711.07414.pdf}.

\bibitem[Hoffman et~al.(2019)Hoffman, Mueller, Klein, and
  Litman]{hoffman2019metricsexplainableaichallenges}
Robert~R. Hoffman, Shane~T. Mueller, Gary Klein, and Jordan Litman.
\newblock Metrics for explainable ai: Challenges and prospects, 2019.
\newblock URL \url{https://arxiv.org/abs/1812.04608}.

\bibitem[Honovich et~al.(2022)Honovich, Shaham, Bowman, and
  Levy]{honovich2022instruction}
Or~Honovich, Uri Shaham, Samuel~R Bowman, and Omer Levy.
\newblock Instruction induction: From few examples to natural language task
  descriptions.
\newblock \emph{arXiv preprint arXiv:2205.10782}, 2022.

\bibitem[Jacovi \& Goldberg(2020)Jacovi and
  Goldberg]{jacovi-goldberg-2020-towards}
Alon Jacovi and Yoav Goldberg.
\newblock Towards faithfully interpretable {NLP} systems: How should we define
  and evaluate faithfulness?
\newblock In \emph{Proceedings of the Association for Computational
  Linguistics}, 2020.
\newblock URL \url{https://aclanthology.org/2020.acl-main.386}.

\bibitem[Kambhampati et~al.(2025)Kambhampati, Stechly, Valmeekam, Saldyt,
  Bhambri, Palod, Gundawar, Samineni, Kalwar, and Biswas]{kambhampati2025stop}
Subbarao Kambhampati, Kaya Stechly, Karthik Valmeekam, Lucas Saldyt, Siddhant
  Bhambri, Vardhan Palod, Atharva Gundawar, Soumya~Rani Samineni, Durgesh
  Kalwar, and Upasana Biswas.
\newblock Stop anthropomorphizing intermediate tokens as reasoning/thinking
  traces!
\newblock \emph{arXiv preprint arXiv:2504.09762}, 2025.

\bibitem[Kasneci et~al.(2023)Kasneci, Se{\ss}ler, K{\"u}chemann, Bannert,
  Dementieva, Fischer, Gasser, Groh, G{\"u}nnemann, H{\"u}llermeier,
  et~al.]{kasneci2023chatgpt}
Enkelejda Kasneci, Kathrin Se{\ss}ler, Stefan K{\"u}chemann, Maria Bannert,
  Daryna Dementieva, Frank Fischer, Urs Gasser, Georg Groh, Stephan
  G{\"u}nnemann, Eyke H{\"u}llermeier, et~al.
\newblock Chatgpt for good? on opportunities and challenges of large language
  models for education.
\newblock \emph{Learning and individual differences}, 103:\penalty0 102274,
  2023.

\bibitem[Khandekar et~al.(2024)Khandekar, Jin, Xiong, Dunn, Applebaum, Anwar,
  Sarfo-Gyamfi, Safranek, Anwar, Zhang, et~al.]{khandekar2024medcalc}
Nikhil Khandekar, Qiao Jin, Guangzhi Xiong, Soren Dunn, Serina Applebaum, Zain
  Anwar, Maame Sarfo-Gyamfi, Conrad Safranek, Abid Anwar, Andrew Zhang, et~al.
\newblock Medcalc-bench: Evaluating large language models for medical
  calculations.
\newblock \emph{Advances in Neural Information Processing Systems},
  37:\penalty0 84730--84745, 2024.

\bibitem[Korbak et~al.(2025)Korbak, Balesni, Barnes, Bengio, Benton, Bloom,
  Chen, Cooney, Dafoe, Dragan, Emmons, Evans, Farhi, Greenblatt, Hendrycks,
  Hobbhahn, Hubinger, Irving, Jenner, Kokotajlo, Krakovna, Legg, Lindner, Luan,
  Mądry, Michael, Nanda, Orr, Pachocki, Perez, Phuong, Roger, Saxe, Shlegeris,
  Soto, Steinberger, Wang, Zaremba, Baker, Shah, and
  Mikulik]{korbak2025chainthoughtmonitorabilitynew}
Tomek Korbak, Mikita Balesni, Elizabeth Barnes, Yoshua Bengio, Joe Benton,
  Joseph Bloom, Mark Chen, Alan Cooney, Allan Dafoe, Anca Dragan, Scott Emmons,
  Owain Evans, David Farhi, Ryan Greenblatt, Dan Hendrycks, Marius Hobbhahn,
  Evan Hubinger, Geoffrey Irving, Erik Jenner, Daniel Kokotajlo, Victoria
  Krakovna, Shane Legg, David Lindner, David Luan, Aleksander Mądry, Julian
  Michael, Neel Nanda, Dave Orr, Jakub Pachocki, Ethan Perez, Mary Phuong,
  Fabien Roger, Joshua Saxe, Buck Shlegeris, Martín Soto, Eric Steinberger,
  Jasmine Wang, Wojciech Zaremba, Bowen Baker, Rohin Shah, and Vlad Mikulik.
\newblock Chain of thought monitorability: A new and fragile opportunity for ai
  safety, 2025.
\newblock URL \url{https://arxiv.org/abs/2507.11473}.

\bibitem[Lage et~al.(2019)Lage, Chen, He, Narayanan, Kim, Gershman, and
  Doshi-Velez]{lage2019evaluation}
Isaac Lage, Emily Chen, Jeffrey He, Menaka Narayanan, Been Kim, Sam Gershman,
  and Finale Doshi-Velez.
\newblock An evaluation of the human-interpretability of explanation.
\newblock \emph{ArXiv}, 2019.
\newblock URL \url{https://arxiv.org/pdf/1902.00006.pdf}.

\bibitem[Lanham et~al.(2023)Lanham, Chen, Radhakrishnan, Steiner, Denison,
  Hernandez, Li, Durmus, Hubinger, Kernion, et~al.]{lanham2023measuring}
Tamera Lanham, Anna Chen, Ansh Radhakrishnan, Benoit Steiner, Carson Denison,
  Danny Hernandez, Dustin Li, Esin Durmus, Evan Hubinger, Jackson Kernion,
  et~al.
\newblock Measuring faithfulness in chain-of-thought reasoning.
\newblock \emph{arXiv preprint arXiv:2307.13702}, 2023.

\bibitem[Levy et~al.(2025)Levy, Elyoseph, and Goldberg]{levy2025humans}
Mosh Levy, Zohar Elyoseph, and Yoav Goldberg.
\newblock Humans perceive wrong narratives from ai reasoning texts.
\newblock \emph{arXiv preprint arXiv:2508.16599}, 2025.

\bibitem[Li et~al.(2023)Li, Shrivastava, Li, Hashimoto, and
  Liang]{li2023benchmarking}
Xiang~Lisa Li, Vaishnavi Shrivastava, Siyan Li, Tatsunori Hashimoto, and Percy
  Liang.
\newblock Benchmarking and improving generator-validator consistency of
  language models.
\newblock \emph{arXiv preprint arXiv:2310.01846}, 2023.

\bibitem[Lin et~al.(2025)Lin, Wu, and
  Sun]{lin2025trainingllmsehrbasedreasoning}
Jiacheng Lin, Zhenbang Wu, and Jimeng Sun.
\newblock Training llms for ehr-based reasoning tasks via reinforcement
  learning, 2025.
\newblock URL \url{https://arxiv.org/abs/2505.24105}.

\bibitem[Liu et~al.(2025)Liu, Diao, Lu, Hu, Dong, Choi, Kautz, and
  Dong]{liu2025prorl}
Mingjie Liu, Shizhe Diao, Ximing Lu, Jian Hu, Xin Dong, Yejin Choi, Jan Kautz,
  and Yi~Dong.
\newblock Prorl: Prolonged reinforcement learning expands reasoning boundaries
  in large language models.
\newblock \emph{arXiv preprint}, 2025.
\newblock URL \url{https://arxiv.org/abs/2505.24864}.

\bibitem[Mancoridis et~al.(2025)Mancoridis, Weeks, Vafa, and
  Mullainathan]{mancoridis2025potemkin}
Marina Mancoridis, Bec Weeks, Keyon Vafa, and Sendhil Mullainathan.
\newblock Potemkin understanding in large language models.
\newblock \emph{arXiv preprint arXiv:2506.21521}, 2025.

\bibitem[Naous et~al.(2025)Naous, Laban, Xu, and Neville]{naous2025flipping}
Tarek Naous, Philippe Laban, Wei Xu, and Jennifer Neville.
\newblock Flipping the dialogue: Training and evaluating user language models.
\newblock \emph{arXiv preprint arXiv:2510.06552}, 2025.

\bibitem[Novikov et~al.(2025)Novikov, V{\~u}, Eisenberger, Dupont, Huang,
  Wagner, Shirobokov, Kozlovskii, Ruiz, Mehrabian,
  et~al.]{novikov2025alphaevolve}
Alexander Novikov, Ng{\^a}n V{\~u}, Marvin Eisenberger, Emilien Dupont, Po-Sen
  Huang, Adam~Zsolt Wagner, Sergey Shirobokov, Borislav Kozlovskii,
  Francisco~JR Ruiz, Abbas Mehrabian, et~al.
\newblock Alphaevolve: A coding agent for scientific and algorithmic discovery.
\newblock \emph{arXiv preprint arXiv:2506.13131}, 2025.

\bibitem[OpenAI(2025)]{openai2025gpt}
OpenAI.
\newblock gpt-oss-120b \& gpt-oss-20b model card, 2025.
\newblock URL \url{https://arxiv.org/abs/2508.10925}.

\bibitem[{OpenAI}(2025)]{openaio4mini}
{OpenAI}.
\newblock Introducing openai o3 and o4-mini, 2025.
\newblock URL \url{https://openai.com/index/introducing-o3-and-o4-mini/}.
\newblock Accessed: Sept 2025.

\bibitem[Pfau et~al.(2024)Pfau, Merrill, and Bowman]{pfau2024lets}
Jacob Pfau, William Merrill, and Samuel~R. Bowman.
\newblock Let{\textquoteright}s think dot by dot: Hidden computation in
  transformer language models.
\newblock In \emph{First Conference on Language Modeling}, 2024.
\newblock URL \url{https://openreview.net/forum?id=NikbrdtYvG}.

\bibitem[Prakash et~al.(2025)Prakash, Shapira, Sharma, Riedl, Belinkov, Shaham,
  Bau, and Geiger]{prakash2025language}
Nikhil Prakash, Natalie Shapira, Arnab~Sen Sharma, Christoph Riedl, Yonatan
  Belinkov, Tamar~Rott Shaham, David Bau, and Atticus Geiger.
\newblock Language models use lookbacks to track beliefs.
\newblock \emph{arXiv preprint arXiv:2505.14685}, 2025.

\bibitem[Qu et~al.(2025)Qu, Singh, Lee, Setlur, Salakhutdinov, Finn, and
  Kumar]{qu2025rlad}
Yuxiao Qu, Anikait Singh, Yoonho Lee, Amrith Setlur, Ruslan Salakhutdinov,
  Chelsea Finn, and Aviral Kumar.
\newblock Rlad: Training llms to discover abstractions for solving reasoning
  problems.
\newblock \emph{arXiv preprint arXiv:2510.02263}, 2025.

\bibitem[Ribeiro et~al.(2016)Ribeiro, Singh, and
  Guestrin]{10.1145/2939672.2939778}
Marco~Tulio Ribeiro, Sameer Singh, and Carlos Guestrin.
\newblock "why should i trust you?": Explaining the predictions of any
  classifier.
\newblock In \emph{Proceedings of the ACM SIGKDD International Conference on
  Knowledge Discovery and Data Mining}, 2016.
\newblock URL \url{https://doi.org/10.1145/2939672.2939778}.

\bibitem[Romera-Paredes et~al.(2024)Romera-Paredes, Barekatain, Novikov, Balog,
  Kumar, Dupont, Ruiz, Ellenberg, Wang, Fawzi, et~al.]{romera2024mathematical}
Bernardino Romera-Paredes, Mohammadamin Barekatain, Alexander Novikov, Matej
  Balog, M~Pawan Kumar, Emilien Dupont, Francisco~JR Ruiz, Jordan~S Ellenberg,
  Pengming Wang, Omar Fawzi, et~al.
\newblock Mathematical discoveries from program search with large language
  models.
\newblock \emph{Nature}, 625\penalty0 (7995):\penalty0 468--475, 2024.

\bibitem[Schut et~al.(2025)Schut, Toma{\v{s}}ev, McGrath, Hassabis, Paquet, and
  Kim]{schut2025bridging}
Lisa Schut, Nenad Toma{\v{s}}ev, Thomas McGrath, Demis Hassabis, Ulrich Paquet,
  and Been Kim.
\newblock Bridging the human--ai knowledge gap through concept discovery and
  transfer in alphazero.
\newblock \emph{Proceedings of the National Academy of Sciences}, 122\penalty0
  (13):\penalty0 e2406675122, 2025.

\bibitem[Shihab et~al.(2025)Shihab, Akter, and
  Sharma]{shihab2025counterfactual}
Ibne~Farabi Shihab, Sanjeda Akter, and Anuj Sharma.
\newblock Counterfactual sensitivity for faithful reasoning in language models.
\newblock \emph{arXiv preprint arXiv:2509.01544}, 2025.

\bibitem[Shojaee et~al.(2025)Shojaee, Mirzadeh, Alizadeh, Horton, Bengio, and
  Farajtabar]{shojaee2025illusion}
Parshin Shojaee, Iman Mirzadeh, Keivan Alizadeh, Maxwell Horton, Samy Bengio,
  and Mehrdad Farajtabar.
\newblock The illusion of thinking: Understanding the strengths and limitations
  of reasoning models via the lens of problem complexity.
\newblock \emph{arXiv preprint arXiv:2506.06941}, 2025.

\bibitem[Singh et~al.(2023)Singh, Morris, Aneja, Rush, and
  Gao]{singh2023explaining}
Chandan Singh, John~X Morris, Jyoti Aneja, Alexander~M Rush, and Jianfeng Gao.
\newblock Explaining data patterns in natural language with language models.
\newblock In \emph{Proceedings of the 6th BlackboxNLP Workshop: Analyzing and
  Interpreting Neural Networks for NLP}, pp.\  31--55, 2023.

\bibitem[Singh et~al.(2024)Singh, Inala, Galley, Caruana, and
  Gao]{singh2024rethinking}
Chandan Singh, Jeevana~Priya Inala, Michel Galley, Rich Caruana, and Jianfeng
  Gao.
\newblock Rethinking interpretability in the era of large language models.
\newblock \emph{arXiv preprint arXiv:2402.01761}, 2024.

\bibitem[Stechly et~al.(2025)Stechly, Valmeekam, Gundawar, Palod, and
  Kambhampati]{stechly2025beyond}
Kaya Stechly, Karthik Valmeekam, Atharva Gundawar, Vardhan Palod, and Subbarao
  Kambhampati.
\newblock Beyond semantics: The unreasonable effectiveness of reasonless
  intermediate tokens.
\newblock \emph{arXiv preprint arXiv:2505.13775}, 2025.

\bibitem[Sun et~al.()Sun, Yan, and Weng]{sun2025relif}
Chung-En Sun, Ge~Yan, and Tsui-Wei Weng.
\newblock Relif: A reliable, interpretable, and faithful lrm for trustworthy
  reasoning.
\newblock In \emph{Mechanistic Interpretability Workshop at NeurIPS 2025}.

\bibitem[Team(2025)]{qwq32b}
Qwen Team.
\newblock Qwq-32b: Embracing the power of reinforcement learning, March 2025.
\newblock URL \url{https://qwenlm.github.io/blog/qwq-32b/}.

\bibitem[Tekin et~al.(2024)Tekin, Ilhan, Huang, Hu, and Liu]{tekin2024llm}
Selim~Furkan Tekin, Fatih Ilhan, Tiansheng Huang, Sihao Hu, and Ling Liu.
\newblock Llm-topla: Efficient llm ensemble by maximising diversity.
\newblock \emph{arXiv preprint arXiv:2410.03953}, 2024.

\bibitem[Turpin et~al.(2023)Turpin, Michael, Perez, and
  Bowman]{turpin2023language}
Miles Turpin, Julian Michael, Ethan Perez, and Samuel~R Bowman.
\newblock Language models don't always say what they think: Unfaithful
  explanations in chain-of-thought prompting.
\newblock \emph{ArXiv}, 2023.
\newblock URL \url{https://arxiv.org/pdf/2305.04388.pdf}.

\bibitem[Venkatraman et~al.(2025)Venkatraman, Jain, Mittal, Shah, Obando-Ceron,
  Bengio, Bartoldson, Kailkhura, Lajoie, Berseth,
  et~al.]{venkatraman2025recursive}
Siddarth Venkatraman, Vineet Jain, Sarthak Mittal, Vedant Shah, Johan
  Obando-Ceron, Yoshua Bengio, Brian~R Bartoldson, Bhavya Kailkhura, Guillaume
  Lajoie, Glen Berseth, et~al.
\newblock Recursive self-aggregation unlocks deep thinking in large language
  models.
\newblock \emph{arXiv preprint arXiv:2509.26626}, 2025.

\bibitem[Wang et~al.(2023)Wang, Fu, Du, Gao, Huang, Liu, Chandak, Liu,
  Van~Katwyk, Deac, et~al.]{wang2023scientific}
Hanchen Wang, Tianfan Fu, Yuanqi Du, Wenhao Gao, Kexin Huang, Ziming Liu, Payal
  Chandak, Shengchao Liu, Peter Van~Katwyk, Andreea Deac, et~al.
\newblock Scientific discovery in the age of artificial intelligence.
\newblock \emph{Nature}, 620\penalty0 (7972):\penalty0 47--60, 2023.

\bibitem[Wei et~al.(2022)Wei, Wang, Schuurmans, Bosma, Xia, Chi, Le, Zhou,
  et~al.]{wei2022chain}
Jason Wei, Xuezhi Wang, Dale Schuurmans, Maarten Bosma, Fei Xia, Ed~Chi, Quoc~V
  Le, Denny Zhou, et~al.
\newblock Chain-of-thought prompting elicits reasoning in large language
  models.
\newblock \emph{Advances in Neural Information Processing Systems},
  35:\penalty0 24824--24837, 2022.

\bibitem[Wolf et~al.(2020)Wolf, Debut, Sanh, Chaumond, Delangue, Moi, Cistac,
  Rault, Louf, Funtowicz, Davison, Shleifer, von Platen, Ma, Jernite, Plu, Xu,
  Le~Scao, Gugger, Drame, Lhoest, and Rush]{wolf-etal-2020-transformers}
Thomas Wolf, Lysandre Debut, Victor Sanh, Julien Chaumond, Clement Delangue,
  Anthony Moi, Pierric Cistac, Tim Rault, Remi Louf, Morgan Funtowicz, Joe
  Davison, Sam Shleifer, Patrick von Platen, Clara Ma, Yacine Jernite, Julien
  Plu, Canwen Xu, Teven Le~Scao, Sylvain Gugger, Mariama Drame, Quentin Lhoest,
  and Alexander Rush.
\newblock Transformers: State-of-the-art natural language processing.
\newblock In Qun Liu and David Schlangen (eds.), \emph{Proceedings of the 2020
  Conference on Empirical Methods in Natural Language Processing: System
  Demonstrations}, pp.\  38--45, Online, October 2020. Association for
  Computational Linguistics.
\newblock \doi{10.18653/v1/2020.emnlp-demos.6}.
\newblock URL \url{https://aclanthology.org/2020.emnlp-demos.6/}.

\bibitem[Xiong et~al.(2025)Xiong, Chen, Qi, and Lakkaraju]{xiong2025measuring}
Zidi Xiong, Shan Chen, Zhenting Qi, and Himabindu Lakkaraju.
\newblock Measuring the faithfulness of thinking drafts in large reasoning
  models.
\newblock \emph{arXiv preprint arXiv:2505.13774}, 2025.

\bibitem[Ye \& Durrett(2022)Ye and Durrett]{ye-durrett-2022-explanations}
Xi~Ye and Greg Durrett.
\newblock Can explanations be useful for calibrating black box models?
\newblock In \emph{Proceedings of the Association for Computational
  Linguistics}, May 2022.
\newblock URL \url{https://aclanthology.org/2022.acl-long.429}.

\bibitem[Yeo et~al.(2024)Yeo, Satapathy, Goh, and
  Cambria]{yeo2024interpretable}
Wei~Jie Yeo, Ranjan Satapathy, Rick Siow~Mong Goh, and Erik Cambria.
\newblock How interpretable are reasoning explanations from prompting large
  language models?
\newblock \emph{arXiv preprint arXiv:2402.11863}, 2024.

\bibitem[Yu et~al.(2025)Yu, Zhang, Zhu, Yuan, Zuo, Yue, Fan, Liu, Liu, Liu,
  Lin, Lin, Ma, Sheng, Tong, Zhang, Zhang, Zhang, Zhu, Zhu, Chen, Chen, Wang,
  Yu, Dai, Song, Wei, Zhou, Liu, Ma, Zhang, Yan, Qiao, Wu, and
  Wang]{yu2025dapoopensourcellmreinforcement}
Qiying Yu, Zheng Zhang, Ruofei Zhu, Yufeng Yuan, Xiaochen Zuo, Yu~Yue, Tiantian
  Fan, Gaohong Liu, Lingjun Liu, Xin Liu, Haibin Lin, Zhiqi Lin, Bole Ma,
  Guangming Sheng, Yuxuan Tong, Chi Zhang, Mofan Zhang, Wang Zhang, Hang Zhu,
  Jinhua Zhu, Jiaze Chen, Jiangjie Chen, Chengyi Wang, Hongli Yu, Weinan Dai,
  Yuxuan Song, Xiangpeng Wei, Hao Zhou, Jingjing Liu, Wei-Ying Ma, Ya-Qin
  Zhang, Lin Yan, Mu~Qiao, Yonghui Wu, and Mingxuan Wang.
\newblock Dapo: An open-source llm reinforcement learning system at scale,
  2025.
\newblock URL \url{https://arxiv.org/abs/2503.14476}.

\bibitem[Zaman \& Srivastava(2025)Zaman and Srivastava]{zaman2025causal}
Kerem Zaman and Shashank Srivastava.
\newblock A causal lens for evaluating faithfulness metrics.
\newblock \emph{arXiv preprint arXiv:2502.18848}, 2025.

\bibitem[Zhang et~al.(2020)Zhang, Kishore, Wu, Weinberger, and
  Artzi]{zhang2019bertscore}
Tianyi Zhang, Varsha Kishore, Felix Wu, Kilian~Q. Weinberger, and Yoav Artzi.
\newblock Bertscore: Evaluating text generation with {BERT}.
\newblock In \emph{8th International Conference on Learning Representations,
  {ICLR} 2020, Addis Ababa, Ethiopia, April 26-30, 2020}. OpenReview.net, 2020.
\newblock URL \url{https://openreview.net/forum?id=SkeHuCVFDr}.

\bibitem[Zhao et~al.(2025)Zhao, Tan, Ma, Li, Jiang, Wang, Yang, and
  Liu]{zhao2025chain}
Chengshuai Zhao, Zhen Tan, Pingchuan Ma, Dawei Li, Bohan Jiang, Yancheng Wang,
  Yingzhen Yang, and Huan Liu.
\newblock Is chain-of-thought reasoning of llms a mirage? a data distribution
  lens.
\newblock \emph{arXiv preprint arXiv:2508.01191}, 2025.

\bibitem[Zhong et~al.(2023)Zhong, Zhang, Li, Ahn, Klein, and
  Steinhardt]{zhong2023goaldd}
Ruiqi Zhong, Peter Zhang, Steve Li, Jinwoo Ahn, Dan Klein, and Jacob
  Steinhardt.
\newblock Goal driven discovery of distributional differences via language
  descriptions.
\newblock 2023.

\bibitem[Zhou et~al.(2021)Zhou, Gandomi, Chen, and
  Holzinger]{zhou2021evaluating}
Jianlong Zhou, Amir~H Gandomi, Fang Chen, and Andreas Holzinger.
\newblock Evaluating the quality of machine learning explanations: A survey on
  methods and metrics.
\newblock \emph{Electronics}, 10\penalty0 (5):\penalty0 593, 2021.

\end{thebibliography}
    \bibliographystyle{iclr2026_conference}
}
\appendix
\clearpage
\appendix

\setcounter{table}{0}
\setcounter{figure}{0}
\renewcommand{\thetable}{A\arabic{table}}
\renewcommand{\thefigure}{A\arabic{figure}}
\renewcommand{\theHfigure}{AppendixFigure\arabic{figure}}
\renewcommand{\theHtable}{AppendixTable\arabic{table}}

\section{Appendix}\label{sec:sample_original_distr}
\subsection{MedCalc Bench data details}
\begin{figure}[!hbt]
    \centering
    \includegraphics[width=\linewidth]{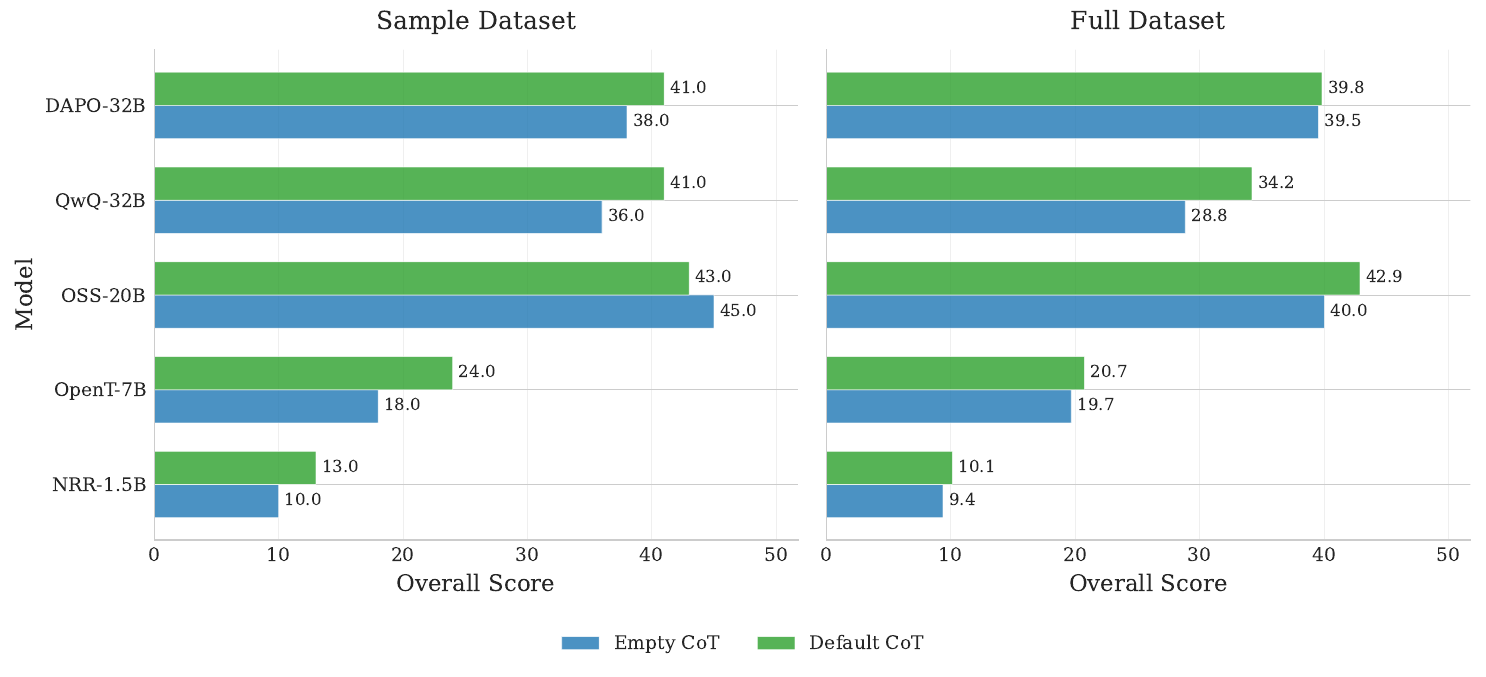}
    \caption{Model performance across all sampled and full data. The trends remain the same. Hence, the sample collected is a representative sample.}
    \label{fig:medcalc-sample-vs-full}
\end{figure}

We randomly sampled 100 data points from the \texttt{MedCalc-Bench} with seed 42 for our experiments. To show that these points are representative, we calculated the default deterministic CoT model performance across the sampled and full dataset.~\Cref{fig:medcalc-sample-vs-full} shows that the trend of best to worst model performance remains the same across default and across empty variations.

The model, \texttt{openai/gpt-oss-20b}, has three reasoning levels --- low, medium, high. Based on the Medcalc benchmark, this model performs the best in the low level reasoning. Hence, for the rest of our experiments, we evaluated this model's CoT in low reasoning level effort.

\subsection{Details on removing answers from explanations}\label{sec:remove_answer_method}
\lstset{
    language=Python,
    basicstyle=\ttfamily\small,
    breaklines=true,
    keywordstyle=\color{blue}\bfseries,
    stringstyle=\color{red},
    commentstyle=\color{green!50!black},
    numberstyle=\tiny\color{gray},
    identifierstyle=\color{black},
    backgroundcolor=\color{gray!10},
    frame=single,
    rulecolor=\color{gray!30},
    showstringspaces=false,
    tabsize=4,
    captionpos=b
}

We prompt OpenAI's o4-mini~\citep{openaio4mini} with the content presented in Listing~\ref{lst:remove_ans_prompt}.

\begin{lstlisting}[caption={Prompt for removing answer from an explanation.}, label={lst:remove_ans_prompt}]
f"""Task: Keep only the hints from the text and remove answer sentences.

Definition: 
- A "hint/explanation sentence" provides guidance that helps someone 
  think about the problem without giving the final solution.  
- An "answer sentence" directly states the final answer, solution, 
  result, or conclusion.

Instructions:  
1. Keep every hint/explanation sentence exactly as written.
2. Remove all answer sentences and statements.
3. Preserve the original wording, order, and formatting of the 
   remaining text.
4. Do not add, rephrase, or generate any new text beyond what is 
   already in the original.
5. Output only the hints.

Original text:
{chain-of-thought}"""
\end{lstlisting}

\subsection{Instruction Induction added data details}
\label{sec:instruction_induction_data}
To create diverse and potentially complex instructions, we construct 12 new tasks in addition to the 24 tasks in the Instruction Induction Dataset~\citep{honovich2022instruction}. The new tasks can be described as follows:
\begin{enumerate}[leftmargin=*]
    \item Reverse from middle: Locate the center point and reverse the left and right segments
    
    \item Smallest Item Length: Find the shortest item and return its character count

    \item Smallest even number square root: Identify the smallest even number and return its square root

    \item Most vowel return consonant: Find the word with the most vowels and return its consonant count

    \item Detect rhyme and rewrite: Detect rhyme schemes in poetry, then rewrite maintaining the same pattern.

    \item Rank by Protein: Group foods into macronutrient categories and order by descending protein percentage
    \item Translate to English: Recognize what language is being used and convert the main phrases into English
    \item Square of Zodiac Animal: Find the zodiac animal in each list and output the square of its zodiac position
    \item Alternate synonym antonym: Alternate between giving an antonym and synonym of the words in the sentence
    \item Most consonant return vowel: Identify the word with the most consonants and return its vowel count
    \item Identify fewest unique letters and return total letter count: Identify the word with the fewest unique letters and return its total letter count
    \item First Word Alphabetically Return Reverse: Find the word that comes first alphabetically and return it in reverse
\end{enumerate}

\begin{figure}[!hbt]
    \centering
    \includegraphics[width=\linewidth]{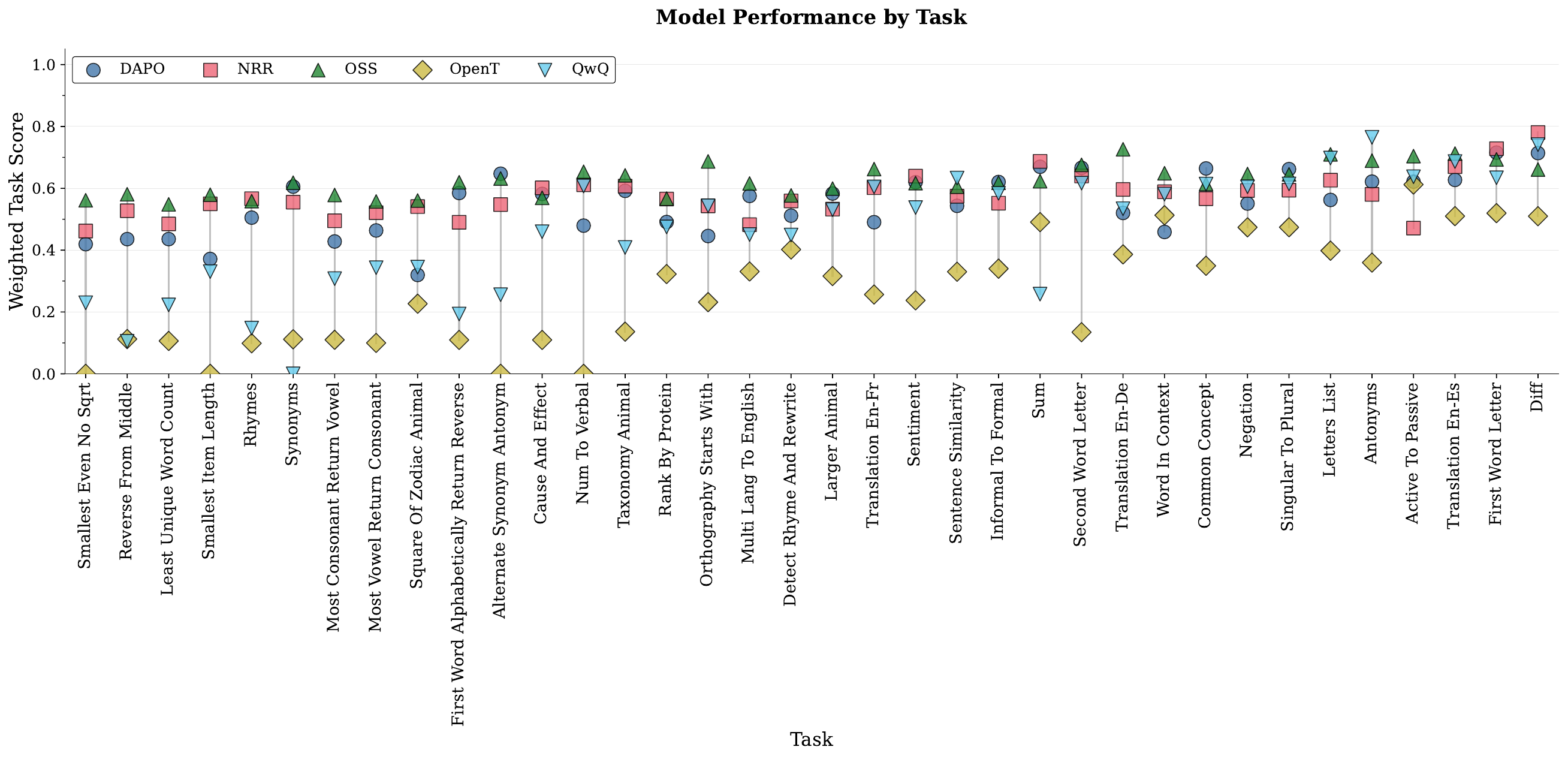}
    \caption{Model performance across all sampled data, including both the original instruction induction and newly introduced tasks.}
    \label{fig:instr-induction-original-all}
\end{figure}

\subsection{Details on ensembled CoT}\label{sec:ensemble_stats}
\begin{figure}[!hbt]
    \centering
    \includegraphics[width=\linewidth]{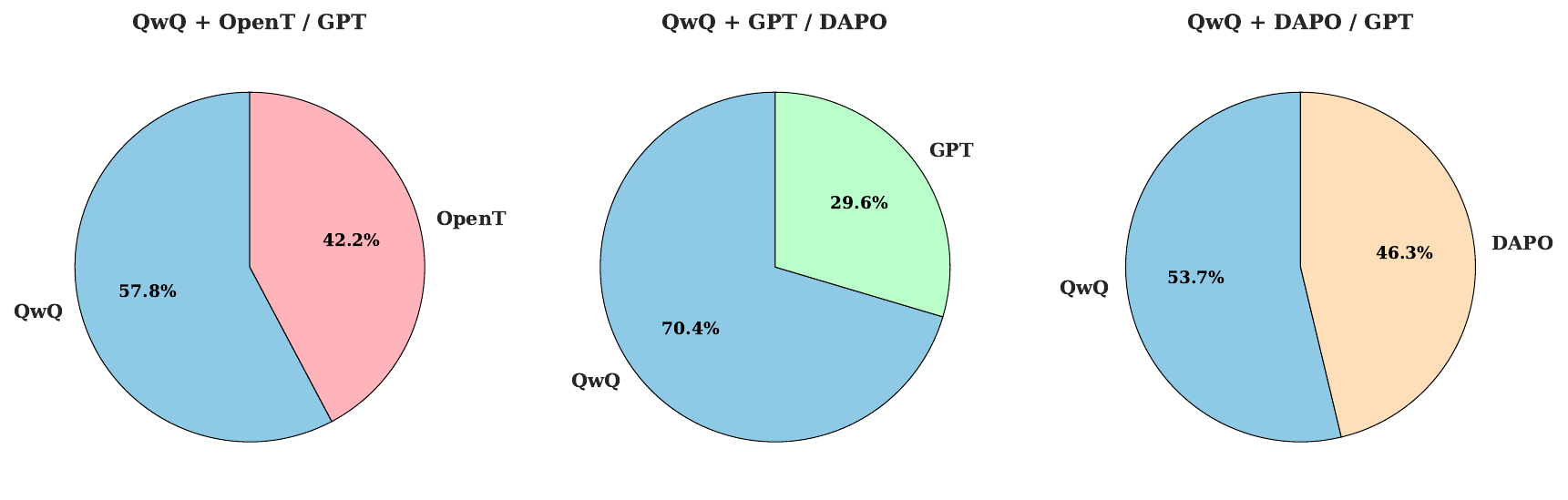}
    \caption{Proportion of selected candidate sentences from different generator models used to construct ensemble chain-of-thoughts across settings in MedCalc-Bench.}
    \label{fig:medcalc-ensemble-model-dist}
\end{figure}

The ensemble chain-of-thought method involves multiple models generating candidate sentences, after which judge or evaluator models select one candidate based on what they are least surprised by seeing, i.e., perplexity. The next sentence is then generated sequentially based on the context of the question and the previous selected candidates, continuing this process iteratively. In ~\Cref{fig:medcalc-ensemble-model-dist}, we look at the distribution of candidates chosen from different pairs of generator models in various combinations of evaluated ensembles.

\begin{figure}[!ht]
    \centering
    \includegraphics[width=0.9\linewidth]{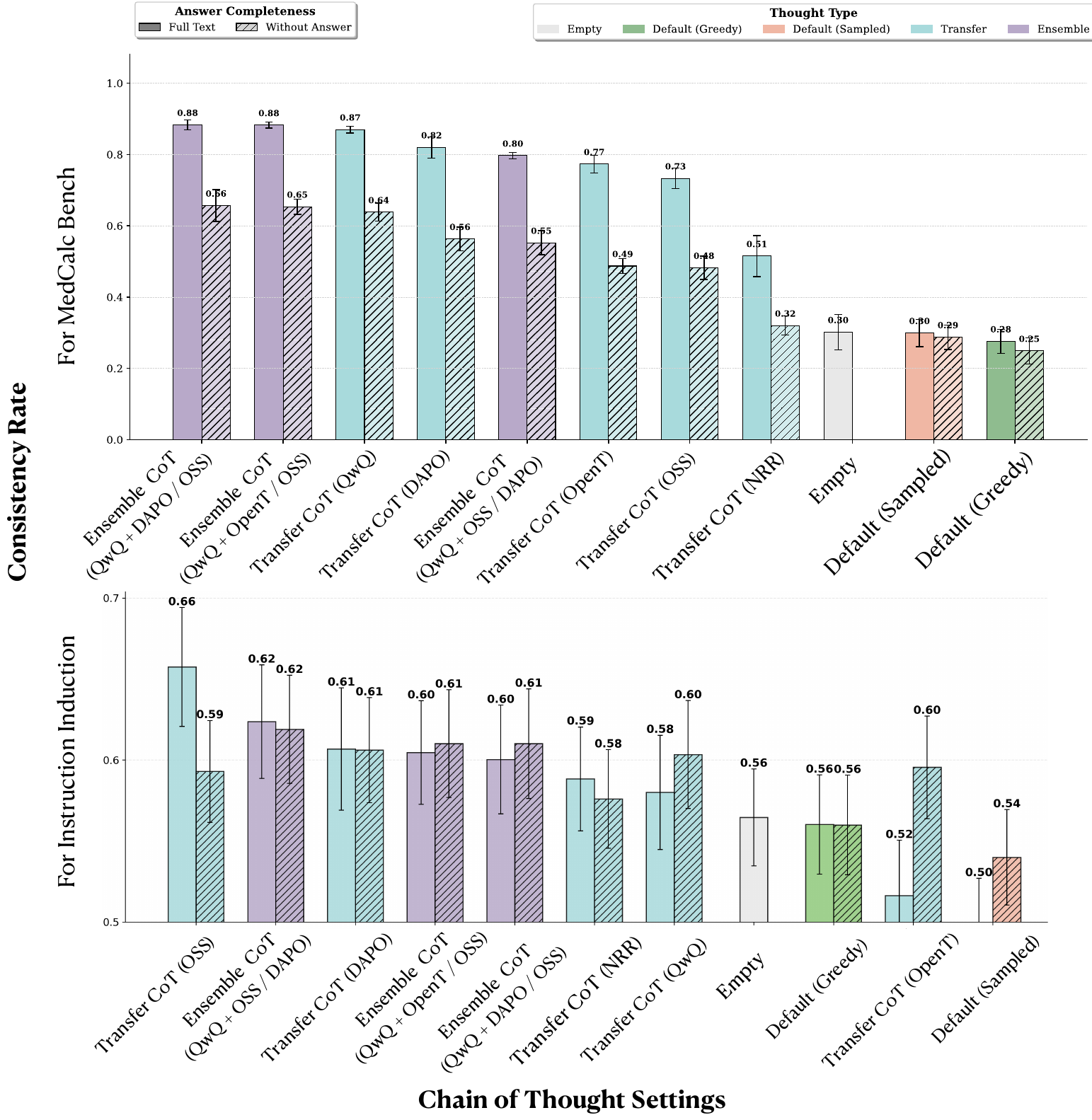}
    \caption{Average pairwise consistency across thought settings in MedCalc-Bench (above) and Instruction Induction (below). For thought variations indicating Ensemble CoT, models listed before the slash (/) serve as generators, while the model after the slash acts as the judge/evaluator.
    Results are reported both for the full text and for text with the final answer removed.
    }
    \label{fig:consistency-all-barplot}
\end{figure}

\subsection{Additional Results}\label{sec:additional_analysis}

\Cref{fig:consistency-all-barplot,tab:accuracy_full} show consistency and accuracy for all settings. ``Full text'' denotes complete model-generated CoTs, while ``w/o answer'' denotes CoTs with direct answers removed, retaining only reasoning steps.

\begin{table}[!htb]
\centering
\footnotesize
\caption{Comparison of CoT accuracy across models for MedCalc-Bench and Instruction Induction.}
\begin{center}
\label{tab:accuracy_full}
\scriptsize
\setlength{\tabcolsep}{3pt}
\begin{tabular}{@{}p{3cm} p{1cm} *{5}{c}  c | *{5}{c} c}
\toprule
\multirow{3}{*}{\textbf{Method}} & \multirow{3}{*}{\textbf{Setting}} & \multicolumn{5}{c}{\textbf{MedCalc-Bench (Exact-Match)}} & \multirow{3}{*}{\textbf{Avg}} & \multicolumn{5}{c}{\textbf{Instruction Induction (BERTScore)}} & \multirow{3}{*}{\textbf{Avg}} \\
\cmidrule(lr){3-7} \cmidrule(lr){9-13}
 & & \textbf{NRR} & \textbf{OpenT} & \textbf{OSS} & \textbf{QwQ} & \textbf{DAPO} & & \textbf{NRR} & \textbf{OpenT} & \textbf{OSS} & \textbf{QwQ} & \textbf{DAPO} & \\
 & & \textbf{1.5B} & \textbf{7B} & \textbf{20B} & \textbf{32B} & \textbf{32B} & & \textbf{1.5B} & \textbf{7B} & \textbf{20B} & \textbf{32B} & \textbf{32B} & \\
\midrule
Empty CoT & No text & 0.10 & 0.18 & 0.45 & 0.36 & 0.38 & 0.29 & 0.53 & 0.55 & 0.56 & 0.55 & 0.57 & 0.55 \\
\midrule
\multirow{2}{*}{Default (Greedy) CoT} 
  & Full text & 0.13 & 0.24 & 0.43 & 0.41 & 0.41 & 0.32 & 0.58 & 0.56 & 0.61 & 0.61 & 0.62 & 0.60  \\
  & W/o Ans & 0.14 & 0.24 & 0.43 & 0.38 & 0.41 & 0.32 & 0.58 & 0.46 & 0.61 & 0.60 & 0.62 & 0.57 \\
\midrule
\multirow{2}{*}{Default (Sampled) CoT} 
  & Full text & 0.14 & 0.32 & 0.47 & 0.39 & 0.37 & 0.34 & 0.58 & 0.50 & 0.52 & 0.57 & 0.60 & 0.55 \\
  & W/o Ans & 0.16 & 0.29 & 0.45 & 0.38 & 0.35 & 0.33 & 0.56 & 0.56 & 0.60 & 0.60 & 0.60 & 0.58 \\
\midrule
\multirow{2}{*}{Trans. CoT (NRR)} 
  & Full text & 0.13 & 0.13 & 0.21 & 0.22 & 0.29 & 0.20 & 0.58 & 0.56 & 0.60 & 0.58 & 0.62 & 0.59 \\
  & W/o Ans & 0.14 & 0.15 & 0.24 & 0.30 & 0.34 & 0.23 & 0.58 & 0.57 & 0.60 & 0.60 & 0.63 & 0.60 \\
\midrule
\multirow{2}{*}{Trans. CoT (OpenT)} 
  & Full text & 0.24 & 0.26 & 0.24 & 0.26 & 0.27 & 0.25 & 0.60 & 0.57 & 0.43 & 0.62 & 0.62 & 0.57 \\
  & W/o Ans & 0.21 & 0.24 & 0.26 & 0.26 & 0.25 & 0.24 & 0.60 & 0.46 & 0.59 & 0.59 & 0.61 & 0.57 \\
\midrule
\multirow{2}{*}{Trans. CoT (OSS)} 
  & Full text & 0.39 & 0.40 & 0.43 & 0.42 & 0.39 & 0.41 & 0.60 & 0.57 & 0.61 & 0.61 & 0.61 & 0.60 \\
  & W/o Ans & 0.26 & 0.44 & 0.43 & 0.40 & 0.44 & 0.39 & 0.60 & 0.57 & 0.61 & 0.60 & 0.62 & 0.60 \\
\midrule
\multirow{2}{*}{Trans. CoT (QwQ)} 
  & Full text & 0.41 & 0.40 & 0.40 & 0.41 & 0.41 & 0.41 & 0.61 & 0.57 & 0.52 & 0.61 & 0.62 & 0.59 \\
  & W/o Ans & 0.34 & 0.37 & 0.39 & 0.38 & 0.37 & 0.37 & 0.60 & 0.57 & 0.61 & 0.60 & 0.62 & 0.60 \\
\midrule
\multirow{2}{*}{Trans. CoT (DAPO)} 
  & Full text & 0.38 & 0.40 & 0.45 & 0.40 & 0.41 & 0.41 & 0.62 & 0.58 & 0.53 & 0.62 & 0.62 & 0.60 \\
  & W/o Ans & 0.31 & 0.40 & 0.39 & 0.40 & 0.41 & 0.38 & 0.60 & 0.58 & 0.62 & 0.61 & 0.62 & 0.61 \\
\midrule
\multirow{2}{*}{Ens. (QwQ+DAPO/OSS)} 
  & Full text & 0.40 & 0.39 & 0.39 & 0.39 & 0.40 & 0.39 & 0.60 & 0.57 & 0.60 & 0.61 & 0.61 & 0.60 \\
  & W/o Ans & 0.39 & 0.37 & 0.41 & 0.41 & 0.40 & 0.40 & 0.60 & 0.57 & 0.61 & 0.60 & 0.62 & 0.60 \\
\midrule
\multirow{2}{*}{Ens. (QwQ+OSS/DAPO)} 
  & Full text & 0.40 & 0.39 & 0.46 & 0.43 & 0.43 & 0.39 & 0.62 & 0.57 & 0.62 & 0.62 & 0.62 & 0.61 \\
  & W/o Ans & 0.28 & 0.37 & 0.45 & 0.42 & 0.43 & 0.38 & 0.60 & 0.57 & 0.61 & 0.60 & 0.62 & 0.60 \\
\midrule
\multirow{2}{*}{Ens. (QwQ+OpenT/OSS)} 
  & Full text & 0.37 & 0.37 & 0.41 & 0.38 & 0.39 & 0.38 & 0.61 & 0.57 & 0.61 & 0.61 & 0.62 & 0.60 \\
  & W/o Ans & 0.34 & 0.41 & 0.42 & 0.38 & 0.40 & 0.39 & 0.61 & 0.58 & 0.61 & 0.60 & 0.62 & 0.60 \\
\bottomrule
\end{tabular}
\end{center}

\end{table}

\begin{figure}[!ht]
    \centering
    \includegraphics[width=0.9\linewidth]{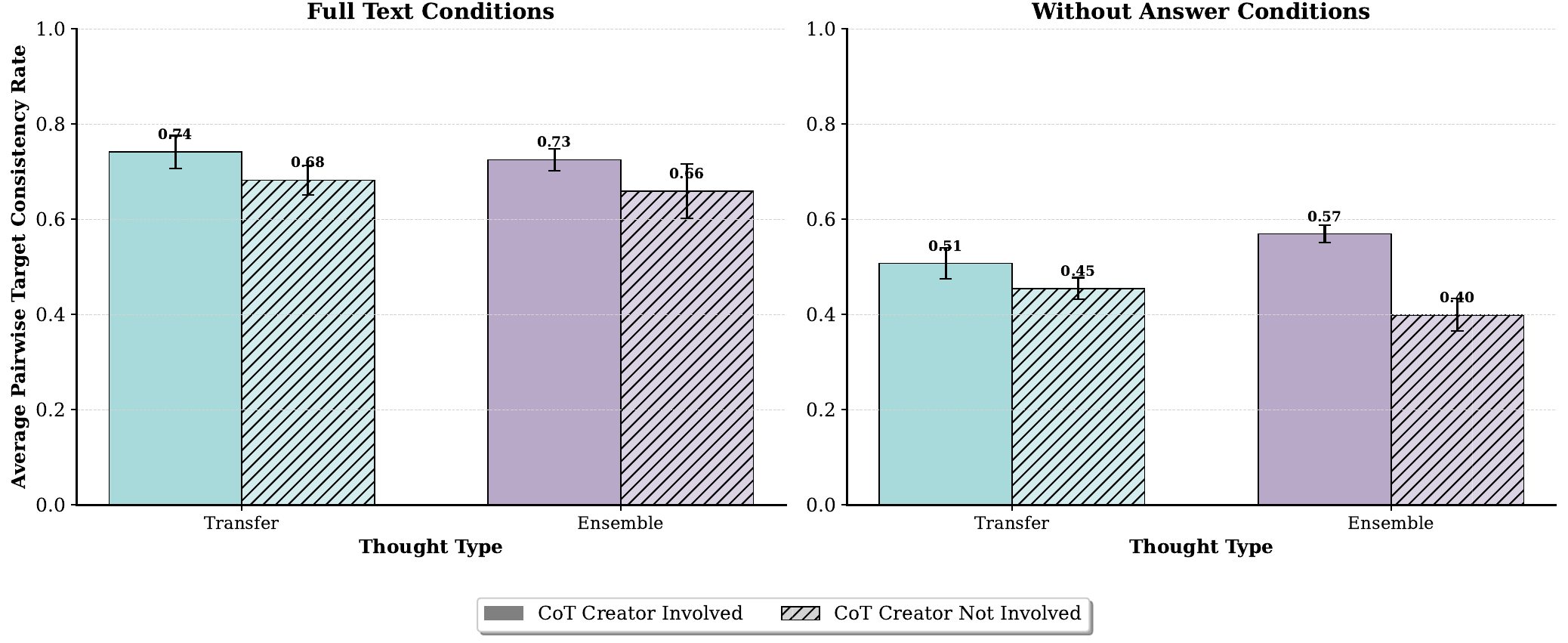}
    \caption{\textbf{LRMs transfer better using their own CoT than the CoT from other models.}
    Plots show the average pairwise consistency for transfer and ensemble thoughts, comparing MedCalc Bench cases where a model is involved as a creator/source of the CoT versus cases where none of the models tested were part of the source.
    }
    \label{fig:model_involvement_consistency}
\end{figure}   

\begin{figure}[!ht]
    \centering
    \includegraphics[width=0.8\linewidth]{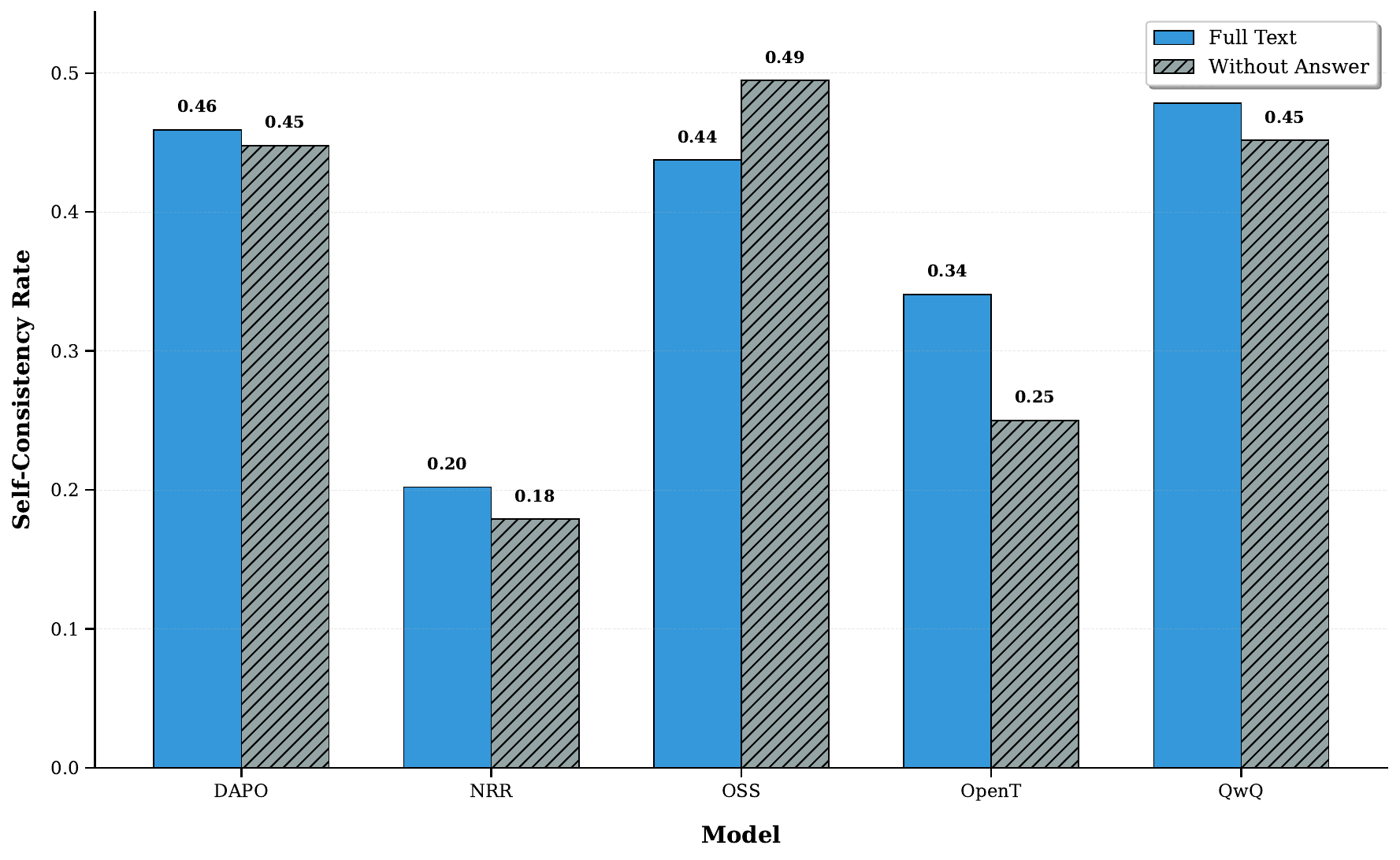}
    \caption{Model self-consistency rates for CoT generation on MedCalc-Bench. Bars compare answer consistency between default (greedy) and sampling-based decoding strategies, with results shown separately for full CoT text (left bar) and CoT with explicit answers removed (right bar).}
    \label{fig:self_consistency}
\end{figure}

\cref{fig:model_involvement_consistency} examines how the average pairwise consistency rate changes depending on whether a pair includes the source model (i.e., the model whose CoT was used) or whether both models are targets that did not contribute to the CoT. This comparison highlights the extent to which similarity in responses persists when one member of the pair is the CoT creator versus when neither model generated the original CoT. \cref{fig:self_consistency} illustrates the models' self-consistency, which ranges from 20\% to about 50\% for respective models in question.
Notably, these self-consistency levels differ from the cross-model consistency observed when models sample their responses.

\subsection{Validation with MedCalc-Bench-Verified}\label{sec:new_version_data_analysis}
Following completion of our main analysis, Khandekar et al.~\cite{khandekar2024medcalc} released MedCalc-Bench-Verified, an updated version with refined ground truth annotations. To validate the robustness of our findings, we re-evaluated our models on the overlapping subset of test samples using the corrected annotations. Our consistency trends remained substantively unchanged, confirming that our results are not artifacts of the original benchmark's annotation quality.

\begin{figure}[!ht]
    \centering
    \includegraphics[width=0.8\linewidth]{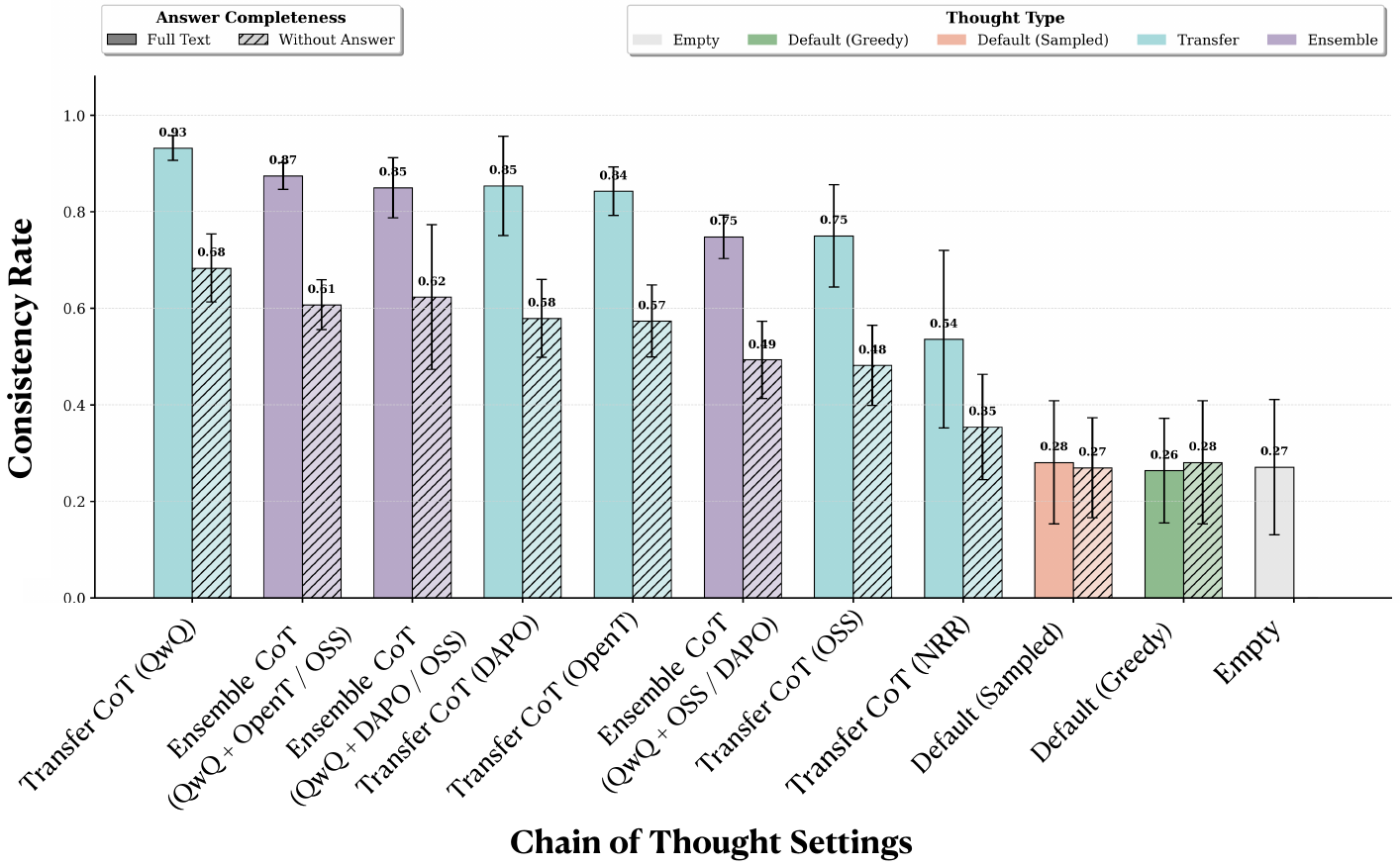}    \caption{Average pairwise consistency across thought settings in MedCalc-Bench-Verified. For thought variations indicating Ensemble CoT, models listed before the slash (/) serve as generators, while the model after the slash acts as the judge/evaluator. Results are reported both for the full text and for text with the final answer removed.}
    \label{fig:medcalc_verified_consistency}
\end{figure}

\subsection{User study details}
\label{sec:user_study_details}
While conducting the user study, no demographic or any other personal identifying information was collected. 

\begin{figure}[!ht]
    \centering
    \includegraphics[width=0.76\linewidth]{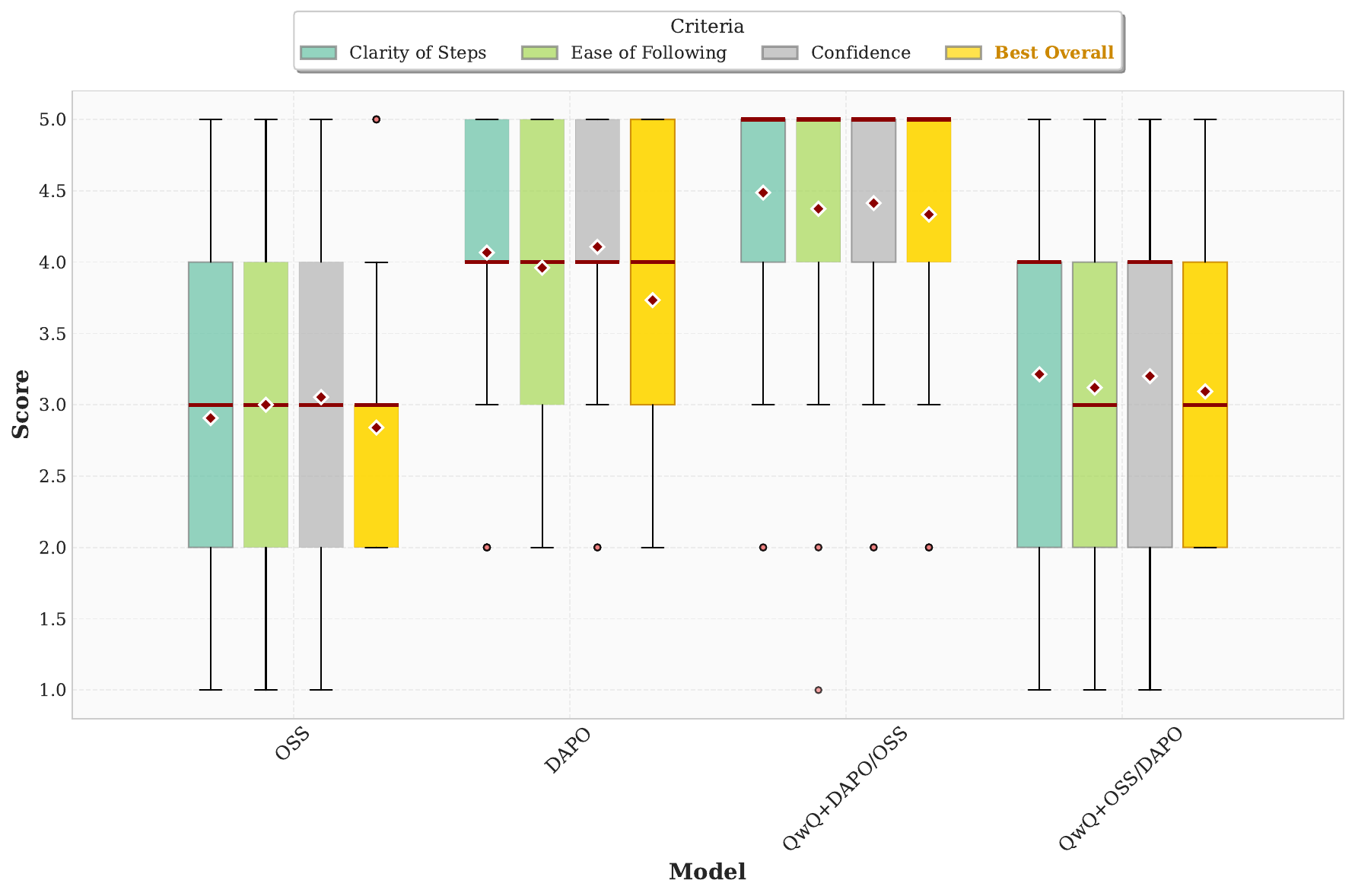}    \caption{\textbf{User study results across evaluation criteria.} Box plots show CoT evaluation scores for each model across four criteria: Clarity of Steps, Ease of Following, Confidence, and Best Overall ranking. Red lines indicate medians, diamonds indicate means. Higher scores are better.}
    \label{fig:user-study-plot-appendix}
\end{figure}

\cref{fig:user-study-plot-appendix} shares a detailed overview of user study scores.
We also share the detailed results of Wilcoxon signed-rank test and
paired t-test in \cref{tab:sig-matrix-complete}.
\begin{table}[!ht]
\centering
\caption{Pairwise Significance Matrix: Mean Differences (Wilcoxon Test)}
\label{tab:sig-matrix-complete}
\small
\begin{tabular}{@{}lcccc@{}}
\toprule
 & \textbf{OSS} & \textbf{DAPO} & \textbf{QwQ+DAPO/OSS} & \textbf{QwQ+OSS/DAPO} \\
\midrule
\multicolumn{5}{l}{\textit{Clarity of Steps}} \\
OSS & -- & \cellcolor{green!20}$-1.16***$ & \cellcolor{green!30}$-1.58***$ & \cellcolor{green!10}$-0.31***$ \\
DAPO & & -- & \cellcolor{green!10}$-0.41***$ & \cellcolor{red!10}$+0.85***$ \\
QwQ+DAPO/OSS & & & -- & \cellcolor{red!20}$+1.26***$ \\
QwQ+OSS/DAPO & & & & -- \\
\midrule
\multicolumn{5}{l}{\textit{Ease of Following}} \\
OSS & -- & \cellcolor{green!20}$-0.96***$ & \cellcolor{green!30}$-1.37***$ & \cellcolor{gray!10}$-0.12$ \\
DAPO & & -- & \cellcolor{green!10}$-0.41***$ & \cellcolor{red!10}$+0.84***$ \\
QwQ+DAPO/OSS & & & -- & \cellcolor{red!20}$+1.25***$ \\
QwQ+OSS/DAPO & & & & -- \\
\midrule
\multicolumn{5}{l}{\textit{Confidence}} \\
OSS & -- & \cellcolor{green!20}$-1.05***$ & \cellcolor{green!30}$-1.36***$ & \cellcolor{green!5}$-0.15*$ \\
DAPO & & -- & \cellcolor{green!10}$-0.31***$ & \cellcolor{red!10}$+0.91***$ \\
QwQ+DAPO/OSS & & & -- & \cellcolor{red!20}$+1.21***$ \\
QwQ+OSS/DAPO & & & & -- \\
\midrule
\multicolumn{5}{l}{\textit{Best Overall}} \\
OSS & -- & \cellcolor{green!20}$-0.89***$ & \cellcolor{green!30}$-1.49***$ & \cellcolor{green!10}$-0.25***$ \\
DAPO & & -- & \cellcolor{green!10}$-0.60***$ & \cellcolor{red!10}$+0.64***$ \\
QwQ+DAPO/OSS & & & -- & \cellcolor{red!20}$+1.24***$ \\
QwQ+OSS/DAPO & & & & -- \\
\bottomrule
\end{tabular}
\begin{tablenotes}
\small
\item Note: Values show mean differences (row model - column model). Green shading indicates row model rated lower (negative difference); red shading indicates row model rated higher (positive difference). Significance: ***$p<0.001$, **$p<0.01$, *$p<0.05$. Gray indicates non-significant difference ($p>0.05$). $n=375$ pairs for all comparisons.
\end{tablenotes}
\end{table}

\end{document}